\pdfoutput=1

\documentclass[11pt]{article}

\usepackage{ACL2023}

\usepackage{times}
\usepackage{latexsym}

\usepackage[T1]{fontenc}

\usepackage[utf8]{inputenc}

\usepackage{microtype}

\usepackage{inconsolata}

\usepackage[dvipsnames]{xcolor}
\usepackage{amsfonts,amsmath,amssymb}

\usepackage{booktabs}
\usepackage{multicol}
\usepackage{multirow}
\usepackage{arydshln}
\usepackage{subcaption}

\usepackage[shortlabels]{enumitem}
\setlist[itemize]{leftmargin=*}
\setlist[enumerate]{leftmargin=*}
\makeatletter
\def\adl@drawiv#1#2#3{%
        \hskip.5\tabcolsep
        \xleaders#3{#2.5\@tempdimb #1{1}#2.5\@tempdimb}%
                #2\z@ plus1fil minus1fil\relax
        \hskip.5\tabcolsep}
\newcommand{\cdashlinelr}[1]{%
  \noalign{\vskip\aboverulesep
           \global\let\@dashdrawstore\adl@draw
           \global\let\adl@draw\adl@drawiv}
  \cdashline{#1}
  \noalign{\global\let\adl@draw\@dashdrawstore
           \vskip\belowrulesep}}
\makeatother

\usepackage{todonotes}

\newcommand{\modelName}{\texttt{OctoBERT}}

\title{World-to-Words: Grounded Open Vocabulary Acquisition \\ through Fast Mapping in Vision-Language Models}

\author{
    Ziqiao Ma\thanks{\enspace Equal contribution.} 
    \hspace{30pt} 
    Jiayi Pan\footnotemark[1] 
    \hspace{30pt} 
    Joyce Chai  \\
    Computer Science and Engineering Division, University of Michigan \\
    \texttt{\{marstin,jiayipan,chaijy\}@umich.edu} 
}

\begin{document}

\maketitle

\begin{abstract}
    
    The ability to connect language units to their referents in the physical world, referred to as {\em grounding}, is crucial to learning and understanding grounded meanings of words. 
    While humans demonstrate fast mapping in new word learning, it remains unclear whether modern vision-language models can truly represent language with their grounded meanings, and how grounding may further bootstrap new word learning.
    To this end, we introduce Grounded Open Vocabulary Acquisition (\texttt{GOVA}) to examine grounding and bootstrapping in open-world language learning.
    As an initial attempt, we propose object-oriented BERT (\texttt{OctoBERT}), a novel visually-grounded language model by pre-training on image-text pairs highlighting grounding as an objective. 
    Through extensive experiments and analysis, we demonstrate that \texttt{OctoBERT} is a more coherent and fast grounded word learner, and that the grounding ability acquired during pre-training helps the model to learn unseen words more rapidly and robustly.\footnote{Code available at \url{https://github.com/sled-group/world-to-words}.}
    
\end{abstract}

\section{Introduction}
\label{sec:intro}

\begin{figure}[!t]
    \centering
    \includegraphics[width=1.0\linewidth]{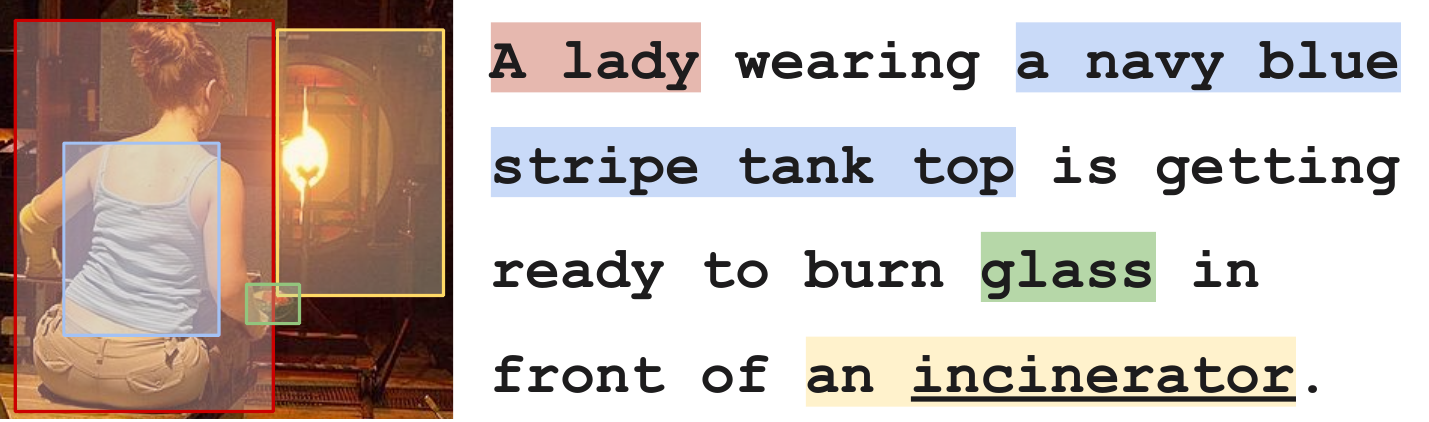}
    \vspace{-15pt}
    \caption{Even when the term ``\texttt{incinerator}'' (highlighted yellow) is new to human learners, they can still locate the most likely referent (indicated by the yellow bounding box) in the perceived world by grounding. }
    \vspace{-15pt}
    \label{fig::mapping}
\end{figure}

Language is learned through sensorimotor experience in the physical world~\citep{bisk2020experience}.
The ability to connect language units to their referents in the physical world, \textit{i.e.} {\em (referential) grounding}, plays an important role in learning and understanding grounded meanings of words~\citep{harnad1990symbol}. 
As shown in Figure~\ref{fig::mapping}, a human reader would easily ground noun phrases to the corresponding entities captured in the image.
Even when the term ``\texttt{incinerator}'' is new to human learners, they can still locate the object of interest through the language and visual context, and acquire its meaning.
In fact, this ability to bootstrap new word learning with only minimal information, known as {\em fast mapping}, is demonstrated abundantly in cognitive literature on human language acquisition~\citep{carey1978acquiring,carey1978child,golinkoff2000becoming,smith2008infants}.

Recently, there has been a substantial effort on pre-training vision-language models (VLMs)~\citep{du2022survey}. 
Despite the exciting performance of these models on a variety of downstream vision and language tasks, it remains unclear whether these models can truly understand or produce language with their grounded meanings in the perceived world, and how grounding may further bootstrap new word learning.
These questions are of interest from both a scientific and an engineering point of view.
From a scientific perspective, grounding is crucial to language learners, as children attend to intended objects in the environment when producing~\citep{tanenhaus1995integration,meyer1998viewing} and comprehending~\citep{smith2007outside} utterances. 
From an engineering perspective, even with the availability of grounded vision language datasets (image-text pairs with fine-grained word-object mappings)~\citep{plummer2015flickr30k}, the costly grounding annotation can hardly cover the whole vocabulary space during the training time. 
Building upon the pre-trained models, it's important for the agent to have the ability to learn grounded new words in a few shots of raw image-text pairs without word-object mappings.

To this end, we introduce Grounded Open Vocabulary Acquisition (\texttt{GOVA}), a scalable formulation to examine grounding and bootstrapping in open-world language learning.
In this formulation, language learning is a combination of learning to predict a word in a linguistic context as well as learning to ground the word in the physical world. 
Under this formulation, we explore the framework in which the model first acquires the grounding ability during pre-training, and then transfers this ability to learn unseen words without grounding supervision.
As an initial step, we developed object-oriented BERT ({\modelName}), a novel visually grounded language model motivated by recent advances in detection transformers (DETR)~\citep{carion2020end,kamath2021mdetr}. 
Compared to many existing VLMs, {\modelName} performs language modeling upon explicit object representations.
The model first acquires the ability to ground during pre-training, and then transfers this intrinsic ability to learn unseen words when grounded supervision is no longer available.

Our empirical results show that learning to map words to their referents plays a significant role in grounded word acquisition. 
By pre-training with fine-grained word-object mappings, {\modelName} demonstrates stronger performance in learning grounded meanings of words, both seen and unseen, yet with orders of magnitude fewer data compared to other competitive VLM baselines. 
The pre-trained model can further provide a foundation for efficient learning of new grounded words with a few examples.
We further present an in-depth analysis to understand potential predictors of VLMs in word learning, which demonstrates intriguing behaviors in comparison to human language learning.
Our findings will facilitate future work on grounded language learning in an open world.

\section{Grounded Open Vocabulary Acquisition (\texttt{GOVA})}
\label{sec:approach}

We start by introducing the settings of \textit{grounded word acquisition} and \textit{few-shot learning of new words} tasks, which are two key components of the Grounded Open Vocabulary Acquisition (\texttt{GOVA}) task formulation.
We further present a unified evaluation protocol and introduce the dataset we curated for this problem.

\vspace*{-5pt}
\subsection{Grounded Word Acquisition}

\begin{figure}[t!]
    \centering
    \includegraphics[width=1.0\linewidth]{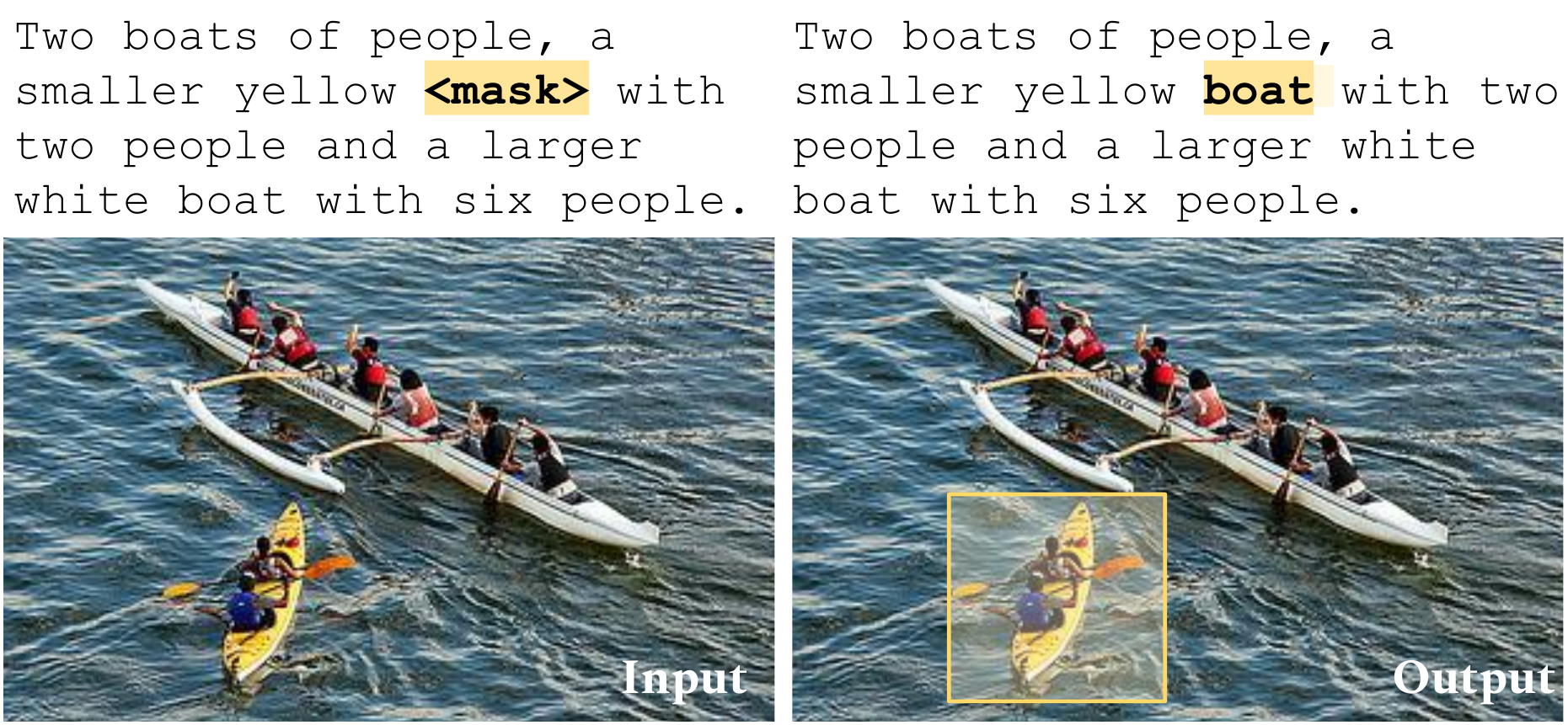}
    \vspace*{-15pt}
    \caption{An instance of the word grounding task. Models are tasked to predict the missing word \texttt{boat} and localize the corresponding smaller yellow boat in the image coherently.}
    \vspace*{-15pt}
    \label{fig::grounding-task}
\end{figure}

Many vision-language tasks have been developed in the past, \textit{e.g.}, visual question answering, visual commonsense reasoning, etc. 
However, these tasks are mainly focused on the end task performance without scrutinizing whether words are grounded to their corresponding visual entities. 
We consider a formulation that directly examines if vision-language models have the ability to acquire grounded meanings of words, specifically, through both \textit{language modeling} and \textit{object localization}.
Figure~\ref{fig::grounding-task} shows an instance of the word acquisition task.
A model is presented with an image $x_{\textrm{img}} \in \mathcal{I}$ and an incomplete caption $x_{\textrm{cap}} \in \mathcal{T}$ with one of its groundable words $w$ (\textit{e.g.}, nouns and adjectives) replaced by a \texttt{MASK}.
The model is tasked to predict this missing word $w \in \mathcal{V}$ based on all available context and localize the corresponding objects $O_w = \{o_1, o_2, \cdots, o_n\}$ in the image by proposing the bounding boxes of them.
Overall, a model capable of solving the grounded word acquisition task is a function $f: \mathcal{I}\times \mathcal{T} \rightarrow \mathcal{V}\times \mathbb{R}^{4n}$.

The language modeling part takes the form of a cloze test, which predicts an open vocabulary word and is widely adopted to evaluate pre-trained language models~\citep{paperno2016lambada,petroni2019language,jin2020visually}.
However, language modeling alone fails to provide a comprehensive evaluation of language grounding.
For example in Figure~\ref{fig::grounding-task}, a model may correctly produce the word ``\texttt{boat},'' but mistakenly attributes the evidence to the larger white boat in the image.
To address this limitation, we require models to localize the corresponding object in the image.
This design is motivated by the disentanglement of object detection into object localization and class recognition~\citep{singh2018r,zareian2021open,zhong2022regionclip}.
It enables vision models to develop a sense of objectness without relying on a predefined set of object classes, thereby potentially allowing them to generalize to unseen objects.
Further comparison with related task setups is discussed in Section~\ref{sec:background} and illustrated in Figure~\ref{appendix::task} in the Appendix.

\vspace*{-5pt}
\subsection{Evaluation Metric}

In language model evaluation, the commonly used measures for assessing performance are the standard hit-rate-at-$k$ (HR$@k$) measure and perplexity~\citep{salazar2020masked,jin2020visually}.
In masked language modeling, the log perplexity of a word $w$ is defined as the log pseudo-perplexity:
\begin{equation}
    \small
    \log\textrm{PPL}(w) = -\log P(w|x_{\textrm{img}},x_{\textrm{cap}})
\end{equation} 
In object detection evaluation, especially for phrase grounding where multiple referents are possible~\citep{kamath2021mdetr}, Any-Protocol and All-Protocol are commonly adopted.
Assuming $n$ ground truth bounding boxes $B = \{b_1, b_2, \cdots, b_n\}$ and $m$ predicted bounding boxes $\widetilde{B} = \{\widetilde{b_1}, \widetilde{b_2}, \cdots, \widetilde{b_m}\}$,
the intersection-over-union (IoU) in both protocols is defined as:
\begin{equation}
\small
    \textrm{IoU}_{\textrm{any}} = \frac{1}{n} \sum_{i\in\{1,2,\cdots,n\}} \max_{j\in\{1,2,\cdots,m\}} \textrm{IoU}(b_i, \widetilde{b_j})
\end{equation}
\vspace*{-0.25cm}
\begin{equation}
\small
    \textrm{IoU}_{\textrm{all}} = \textrm{IoU}(\cup B, \cup \widetilde{B})
\end{equation}

However, these metrics only capture unimodal performance without concerning the correctness of cross-modal mapping. 
We design two new metrics to combine language and vision performance:
\begin{itemize}[leftmargin=*]
    \setlength\itemsep{-0.25em}
    \item {\bf Grounded hit-rate} (G-HR$@k$), the proportion of tests with the masked word appearing in the top-$k$ candidates and a localization IoU over 0.5.
    \item {\bf Grounded perplexity} (G-PPL) as follows:
        \begin{equation}
        \small
            \log\textrm{G-PPL}(w) = 
            \begin{cases} 
            \infty & \mbox{if IoU $ = 0$} \\
            \log\textrm{PPL}(w) - \log \textrm{IoU} & \mbox{else} 
            \end{cases}
        \end{equation}
\end{itemize}

\subsection{Few-Shot Learning of New Words}

\begin{figure}[t!]
    \centering
    \includegraphics[width=1.0\linewidth]{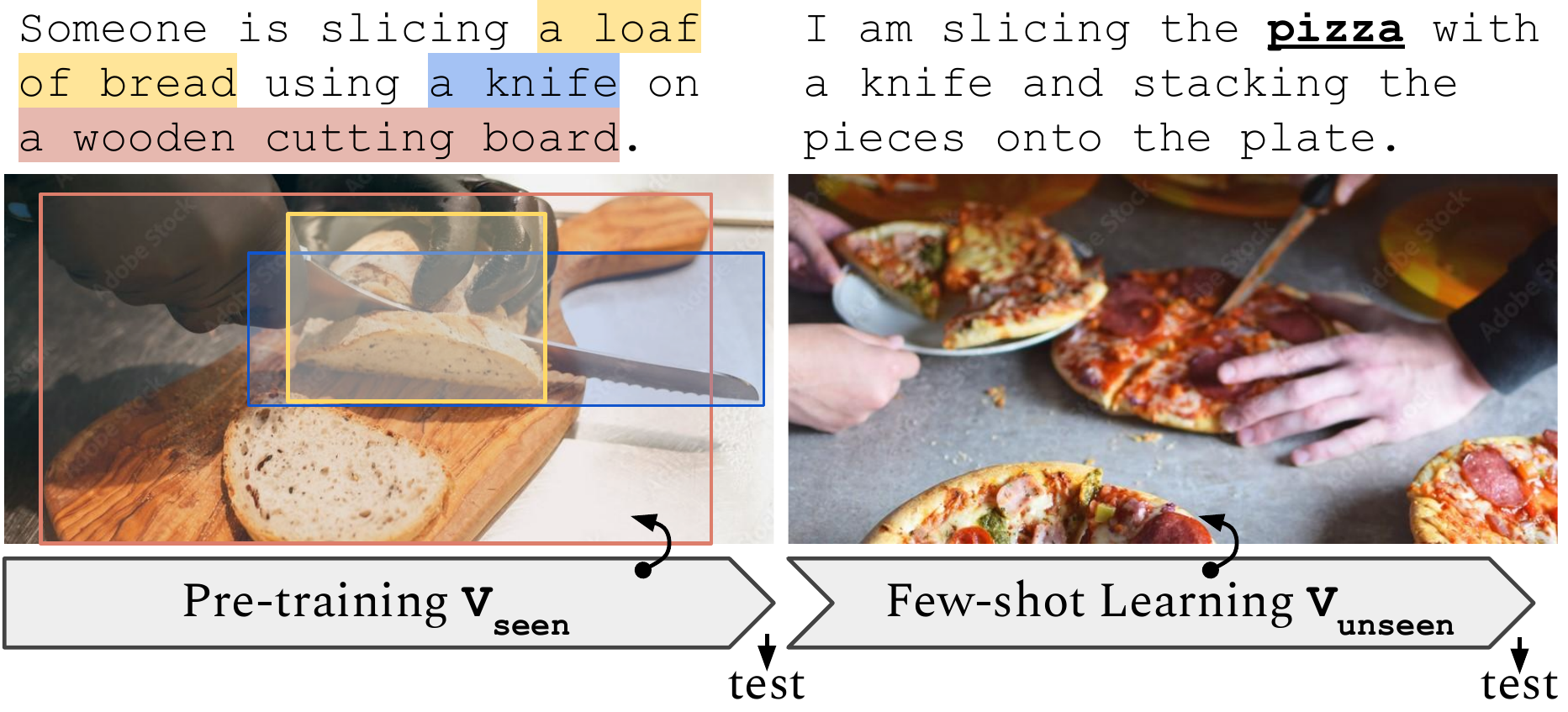}
    \vspace{-20pt}
    \caption{An illustration of the few-shot new word learning paradigm. The model first pre-trains on a grounding dataset with a set of base words ($\mathcal{V}_{\textrm{seen}}$), and then attempts to acquire a set of unseen words ($\mathcal{V}_{\textrm{unseen}}$) in a small number of raw text-image pairs. Tests are performed after each training session. \vspace{-15pt}}
    \label{fig::fewsample}
\end{figure}

Although there are grounding datasets available, \textit{i.e.}, image-text pairs with word-object mapping annotation~\citep{plummer2015flickr30k}, it is impractical to obtain such fine-grained annotation on a large scale and to cover the whole vocabulary space $\mathcal{V}$.
We therefore explore grounded new word learning as a few-shot learning problem, especially under the setting of incremental class learning~\cite{mandziuk1999incremental,kemker2018measuring}.
An intuitive illustration of the few-shot new word learning framework is provided in Figure~\ref{fig::fewsample}. 
Under this framework, a computational model is developed in two stages.
During the pre-training stage, the model receives image-caption pairs, with fine-grained word-object annotation for a set of base words $\mathcal{V}_{\textrm{seen}} \subseteq \mathcal{V}$.
After pre-training, the model is provided with few samples of raw text-image pairs, each containing a set of unseen words $\mathcal{V}_{\textrm{unseen}} \subseteq \mathcal{V}$ that the model has to acquire.

Tests are performed after each training stage.
It's important to note that the unseen words may not be completely new, \textit{e.g.}, the models may have encountered these words in its language encoder initialized with pre-trained language models.
We consider them ``unseen'' because the model never sees these words paired with their referent, \textit{i.e.}, the grounded meanings of the words are unknown.

\begin{figure*}[!htp]
    \centering
    \includegraphics[width=1.0\linewidth]{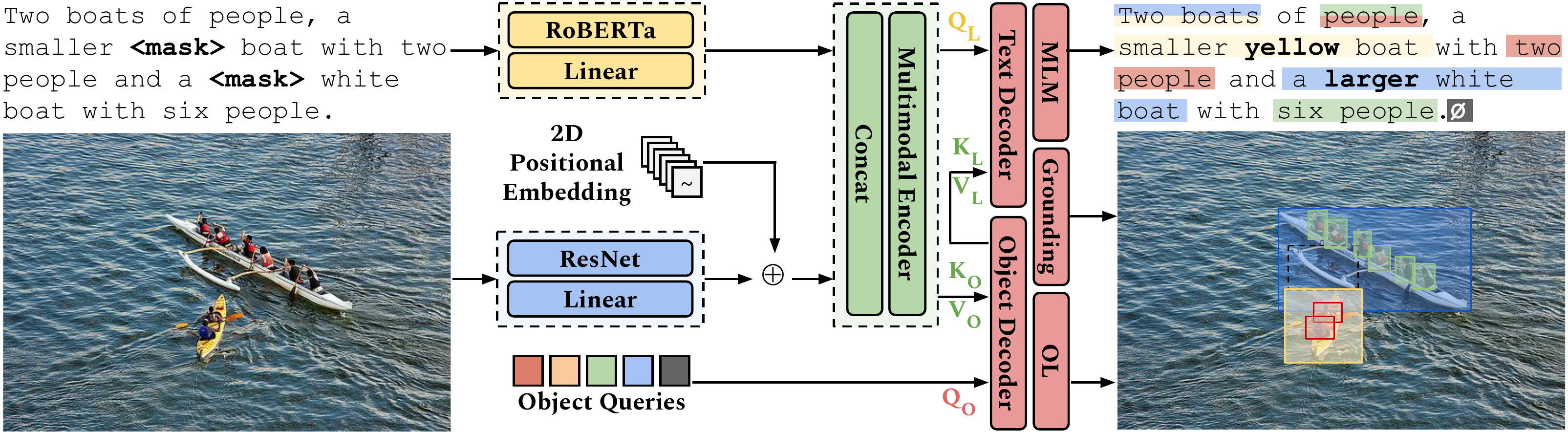}
    \vspace{-17pt}
    \caption{An overview of \texttt{OctoBERT}, a visually grounded language model pre-trained with three objectives: masked language modeling (MLM), object localization (OL), and grounding through word-region alignment (WRA). \vspace{-15pt}}
    \label{fig::model}
\end{figure*}

\subsection{Dataset Curation}

We build our dataset based on the Flickr30K Entities dataset~\citep{plummer2015flickr30k}, which contains image-text pairs with dense annotations between groundable phrases and bounding boxes of objects.
The groundable phrases and regions are defined by the dataset, as chunks of text that refer to object bounding boxes.
To construct word grounding instances, we use Stanza~\citep{qi2020stanza} to parse the caption, enumerate every word in the groundable phrase, and identify those with a POS tag of \texttt{NOUN} or \texttt{ADJ}.
These groundable words are replaced by \texttt{MASK} one at a time and matched to their corresponding bounding boxes.

The dataset is divided into 4 splits: pre-training set, unseen words training set, seen words test set, and unseen words test set.
We start by selecting 31 unseen words and holding out all text-image pairs containing these words from the training split of Flickr30K Entities.
The hold-out text-image pairs are further divided into the training and test sets for unseen words.
The remaining training split of Flickr30K Entities is used for the pre-training set.
To prevent frequent words (\textit{e.g.}, ``\texttt{man}'') from dominating the test results of the seen words, we choose 60 seen words and sample an equal number of test instances for each word from the test split of Flickr30K Entities.
More details and statistics of the dataset are available in Appendix~\ref{appendix:data}.

\section{Computational Models}
\label{sec:model}

\begin{table*}[!htp]
\hspace*{-0.2cm}
\scalebox{0.53}{
    \begin{tabular}{ccccccccccccc}
    \toprule
    \multirow{2}{*}{Models} & \multicolumn{6}{c}{Seen ($|\mathcal{V}_{\text{seen}}|=60$)} &  \multicolumn{6}{c}{Unseen ($|\mathcal{V}_{\text{unseen}}|=31$)}  \\
    \cmidrule(lr){2-7} \cmidrule(lr){8-13}
    & G-HR@1 ($\uparrow$)      & log G-PPL ($\downarrow$)  & HR@1 ($\uparrow$) & log PPL ($\downarrow$) & Acc ($\uparrow$)    & IoU ($\uparrow$) 
    & G-HR@1 ($\uparrow$)      & log G-PPL ($\downarrow$)  & HR@1 ($\uparrow$) & log PPL ($\downarrow$) & Acc ($\uparrow$)    & IoU ($\uparrow$) \\
    \cmidrule(lr){1-1} \cmidrule(lr){2-3} \cmidrule(lr){4-5} \cmidrule(lr){6-7} \cmidrule(lr){8-9} \cmidrule(lr){10-11} \cmidrule(lr){12-13}
    RoBERTa                 & -           & -           & 38.0   & 2.75    & -                    & -                    & -         & -             & 23.1 & 4.96    & -           & -           \\
    RoBERTa (\texttt{FT})            & -           & -           & 47.9 & 1.99    & -                    & -                    & -         & -             & 24.3 & 4.38    & -           & -           \\
    ViLT                    & -           & -           & 64.7 & \textbf{1.27}    & -                    & -                    & -         & -             & 32.7 & 3.68    & -           & -           \\
    MDETR                   & -           & -           & -    & -       & 27.8 / 27.0          & 25.3 / 28.0          & -         & -             & -    & -       & 26.3 / 20.2 & 23.9 / 21.7 \\
    ViLT+MDETR              & 19.8 / 19.3 & 2.53 / 2.43 & 64.7 & \textbf{1.27}    & 31.1 / 30.4          & 28.5 / 31.2          & 8.6 / 8.1 & 5.07 / 5.12   & 32.7 & 3.68    & 27.3 / 23.3 & 25.0 / 23.8 \\
    VisualBERT (\texttt{FT})         & 28.5 / -    & 2.96 / -    & 42.3 & 2.33    & \textbf{68.1} / -             & 53.3 / -             & 10.2 / -  & 5.60 / -      & 20.7 & 4.81    & 50.6 / -    & 45.2 / -    \\
    \cmidrule(lr){1-1} \cmidrule(lr){2-3} \cmidrule(lr){4-5} \cmidrule(lr){6-7} \cmidrule(lr){8-9} \cmidrule(lr){10-11} \cmidrule(lr){12-13}
    {\modelName}$_\textrm{w/o G}$ (\texttt{FT})             & 28.9 / 27.8 & 2.33 / 2.38 & 63.9 & 1.41    & 44.0 / 43.0 & 40.0 / 38.2 & 1.1 / 1.1 & 11.89 / 12.04 & 3.7  & 10.87   & 38.7 / 31.9 & 36.2 / 31.0 \\
    {\modelName}        & \textbf{47.0} / \textbf{46.3} & \textbf{1.79} / \textbf{1.81} & \textbf{66.9} & \textbf{1.26}    & \textbf{66.8} / \textbf{66.3} & \textbf{58.8} / \textbf{57.6} & 2.3 / 2.3 & 11.58 / 11.74 & 4.2  & 11.01   & \textbf{61.3} / \textbf{53.1} & \textbf{56.3} / \textbf{48.0} \\
    \bottomrule
    \end{tabular}
}
\vspace{-5pt}
\caption{Test results on the seen and unseen words, obtained immediately after pre-training. Unless noted explicitly as fine-tuned (\texttt{FT}), all results reflect the performance of models without fine-tuning. Evaluations under both All and Any-protocols are provided in the table as (All/Any) pairs. For models depending on a frozen pre-trained object detector, we can only provide evaluation under All-Protocol. We note that the unseen words are only unseen to {\modelName}, as pre-trained baselines have encountered them during development. We report the results for reference.\vspace{-10pt}}
\label{tab:result}
\end{table*}

\vspace*{-5pt}
\subsection{Object-Oriented BERT (\texttt{\modelName})}

Humans demonstrate fast mapping, the ability to learn new words with only minimal information~\citep{carey1978acquiring,carey1978child,golinkoff2000becoming}.
Motivated by how visual grounding helps humans in bootstrapping new words, we propose a computational framework that first acquires the ability to ground during pre-training, and then transfers this intrinsic ability to learn unseen words when grounded supervision is no longer available.
We introduce object-oriented BERT (\texttt{\modelName}), an end-to-end visually-grounded language model illustrated in Figure~\ref{fig::model}.

\vspace*{-0.1cm}
\paragraph{Model Architecture.} 
Similarly to dual-stream vision-language models, \texttt{\modelName} encodes the textual input with a pre-trained language model~\citep{liu2019roberta}, and encodes image input with convolutional backbone~\citep{he2016deep} with 2D positional encoding added.
The text and image representations are linearly projected onto a joint semantic space and concatenated.
The multimodal representation is then forwarded into a cross-encoder with self-attention layers.
The cross-encoded representations in the final layer are sent into an object decoder, together with a set of learnable object queries. 
The object decoder produces an object embedding for each input object query, which can be considered as a representation of the proposed object.
The object representations are further forwarded to the text decoder, which allows language modeling to explicitly attend to the perceived objects.
We discuss the pre-training objectives, especially how the model acquires grounding in the following paragraphs.
Other details are available in Appendix~\ref{appendix:model}.

\vspace*{-0.1cm}
\paragraph{Masked Language Modeling (MLM).} 
As an intrinsic task, we follow the majority of existing pre-trained vision-language models to perform masked language modeling with a two-layer MLP.
Words in input text are randomly masked out, and the model predicts the masked words conditioned on the corrupted sentence and image.
Words in groundable phrases are masked with a probability of 0.4 and those in non-groundable regions are masked with a lower probability of 0.1.

\vspace*{-0.1cm}
\paragraph{Object Localization (OL).} 
Each object representation will be decoded by a shared three-layer MLP to produce a bounding box.
We follow prior detection transformers (DETR)~\citep{carion2020end,kamath2021mdetr} to perform bipartite matching between proposed boxes and ground truth boxes with a Hungarian loss~\citep{kuhn1955hungarian}.
The predicted boxes are optimized towards ground truth using the generalized intersection-over-union (GIoU) loss~\citep{rezatofighi2019generalized} and the L1 loss.

\vspace*{-0.1cm}
\paragraph{Grounding.} 
The notion of {\em Grounding} is realized by grounded pre-training through word-region alignment (WRA) which enables fine-grained cross-modal mapping between words and objects.
It consists of two levels of alignment: \textit{positional alignment} and \textit{semantic alignment}.
In positional alignment, the model learns to map each object representation to words in the sentence, which could possibly be a \texttt{MASK} or an additional no-object label $\varnothing$~\citep{yu2013grounded,kamath2021mdetr}.
We use a fully-connected layer to predict the distribution over token positions with cross-entropy loss.
In semantic alignment, the model learns to bring word representations closer to the object representations that they ground to, and push the unrelated pairs farther. 
We use a contrastive loss over the final layers of the object and text decoders.

\vspace*{-0.1cm}
\subsection{Baselines}
\label{sec:baselines}

\vspace*{-0.05cm}
\paragraph{Groundless Baseline.}

A baseline with no grounding ability is developed by pre-training \texttt{\modelName} in the same condition but removing the grounding objectives in the loss function.
We refer to this groundless model as \texttt{\modelName}$_\textrm{w/o G}$.
Like a typical pre-trained VLM, \textit{e.g.}, VisualBERT~\citep{li2019visualbert}, \texttt{\modelName}$_\textrm{w/o G}$ performs language modeling based on the object features, without explicit cross-modal referential grounding.
We apply \texttt{\modelName}$_\textrm{w/o G}$ on \texttt{GOVA} task by fine-tuning the model on the pre-training dataset with grounding objective until convergence.

\vspace*{-0.1cm}
\paragraph{Pre-trained Baselines.}
For the majority of the pre-trained VLMs, the unseen words are known during pre-training.
Also, the primary focus of this work is to understand grounding and bootstrapping in grounded word acquisition.  
It's not our goal to scale up or re-train all variants of pre-training frameworks.
Therefore, we compare our model to the pre-trained VLMs with equal or reasonably larger scales for only reference and analysis purposes.
We choose representative baselines in phrase grounding, as presented in Table~\ref{tab:result}:
\begin{itemize}[leftmargin=*]
    \setlength\itemsep{-0.25em}
    \item ``Detect-and-Recognize'' Baseline: Models under this framework rely on a pre-trained frozen object detector, and then learn to predict words from proposed objects. We choose the fine-tuned VisualBERT~\citep{li2019visualbert} for this type.
    \item ``Produce-and-Localize'' Baseline: Models under this framework rely on a pre-trained vision-language model to predict the missing word, and then perform referring expression comprehension and propose objects. We combine ViLT~\citep{kim2021vilt} and MDETR~\citep{kamath2021mdetr} for their competitive performance in vision-conditioned language modeling and phrase grounding individually.
\end{itemize}

\section{Empirical Findings}
\label{sec:experiment}

\subsection{Grounded Pre-training}

The results of this section are obtained from the test immediately following pre-training.

\begin{table}[!htp]
    \vspace{4pt}
    \scalebox{0.705}{
        \begin{tabular}{ccccc}
        \toprule
        Models          & \# Param & \# Imgs & \# Caps & Objectives \\ 
        \cmidrule(lr){1-1} \cmidrule(lr){2-5}
        RoBERTa         & 120M     & -       & -       & MLM \\
        VisualBERT      & 180M     & 200K    & 567K    & MLM, ITM \\
        ViLT            & 110M     & 4.0M    & 10M     & WRA*, MLM, ITM \\
        MDETR           & 200M     & 200K    & 1.3M    & WRA, OL \\
        \cmidrule(lr){1-1} \cmidrule(lr){2-5}
        {\modelName}   & 200M     & 30K     & 150K    & WRA, MLM, OL\\
        {\modelName}$_\textrm{w/o G}$ & 200M     & 30K     & 150K    & MLM, OL \\
        \bottomrule
        \multicolumn{5}{l}{*WRA is formulated as word-patch alignment in ViLT, thus it} \\
        \multicolumn{5}{l}{cannot perform object localization without major modifications.}
        \end{tabular}
    }
    \vspace{-5pt}
    \caption{The baselines for comparisons and references. ITM stands for Image Text Matching, and all the other abbreviations follow Section~\ref{sec:approach}.\vspace{-15pt}}
    \label{tab:baselines}
\end{table}

\vspace{-0.1cm}
\paragraph{Pre-training Results on Seen Words}

The main results for the pre-training stage are summarized in Table~\ref{tab:result}.
Our direct observation is the strong performance of {\modelName} in terms of both grounded metrics, Top-1 Grounded Hit-Rate (G-HR$@1$) and Grounded Perplexity (G-PPL).
{\modelName} significantly outperforms the groundless baseline \texttt{\modelName}$_\textrm{w/o G}$ and pre-trained baselines, even for systems pre-trained with a significantly larger amount of data and computing, as shown in Table~\ref{sec:approach}. 
While {\modelName} produces correct predictions of the missing words as well as the locations of the corresponding bounding boxes, it turns out to be challenging for baselines to achieve them both. 
For ``Detect-and-Recognize'' baseline (VisualBERT), we observe a comparable object localization performance empowered by the frozen object detector.
However, it suffers from a poor language modeling ability (as demonstrated by HR$@1$ and PPL, weaker than a fine-tuned RoBERTa).
For the ``Produce-and-Localize'' baseline (ViLT+MDETR), we observe a strong language modeling performance due to the scale of ViLT.
Yet, correct word grounding remains difficult, as can be seen from the poor localization performance.
These results demonstrate that the \texttt{\texttt{GOVA}} task is challenging, and {\modelName} is competitive in learning grounded word meanings during pre-training.

\vspace{-0.1cm}
\paragraph{Bootstrapping through Grounded Objectives.}

We further provide a cross-time analysis to understand the role of grounded objectives in pre-training efficiency.
The results of different training steps are provided in Table~\ref{tab:ablation}.
From the table, we observe that {\modelName} outperforms both of its groundless variants in language modeling, object localization, and jointly under the grounded perplexity. 
What's even more striking is that {\modelName} achieves better performance with {\em 10 times less training data} compared to the model trained without the grounding objective (\textit{i.e.}, the WRA objective).  
These results confirm the crucial role of explicit word-object alignment in efficient grounded word learning.
This can be explained by that the grounded objectives attempt to align the vision and language semantic spaces, which ideally benefit both visually conditioned language modeling and language-conditioned object localization.
Although it is possible to build a mapping between word and object representations through cross-modal probing and fine-tuning after pre-training, these methods are not comparable to systems with grounded objectives in terms of efficiency and performance.

\begin{table}[!htp]
\scalebox{0.81}{
\begin{tabular}{cccc}
\toprule
\# Steps & Metrics  & {\small\modelName} &  {\small\modelName}$_\textrm{w/o G}$ (\texttt{FT}) \\ 
\cmidrule(lr){1-1} \cmidrule(lr){2-2} \cmidrule(lr){3-4}
\multirow{3}{*}{10k}  & IoU ($\uparrow$)         & \textbf{46.7} / \textbf{46.2}   & 36.9 / 35.3       \\
                     & log PPL ($\downarrow$)   &  \textbf{1.46}           & 1.53 \\
                     & log G-PPL ($\downarrow$) & \textbf{2.22} / \textbf{2.23}   &  2.52 / 2.57       \\
                     \cmidrule(lr){1-1} \cmidrule(lr){2-2} \cmidrule(lr){3-4}
\multirow{3}{*}{50k}  & IoU ($\uparrow$)         & \textbf{58.1} / \textbf{57.1}    & 39.6 / 38.8       \\
                     & log PPL ($\downarrow$)   & \textbf{1.26}        &  1.44             \\
                     & log G-PPL ($\downarrow$) & \textbf{1.80} / \textbf{1.82}    & 2.34 / 2.38       \\
                     \cmidrule(lr){1-1} \cmidrule(lr){2-2} \cmidrule(lr){3-4}
\multirow{3}{*}{100k} & IoU ($\uparrow$)         & \textbf{58.7} / \textbf{57.6}     & 40.0 / 38.2       \\
                     & log PPL ($\downarrow$)   & \textbf{1.26}          & 1.41              \\
                     & log G-PPL ($\downarrow$) & \textbf{1.79} / \textbf{1.81}   & 2.34 / 2.38 \\
                     \bottomrule
\end{tabular}
}
\vspace{-8pt}
\caption{Comparison of {\modelName} and its non-grounding version at different training steps. {\modelName}$_\textrm{w/o G}$ is evaluated with fine-tuning. Both Any and All-protocols are provided in the table as (All/Any). \vspace{-20pt}}
\label{tab:ablation}
\end{table}

\vspace{-0.1cm}
\paragraph{Pre-training Results on Unseen Words: Word-Agnostic Grounding}

\begin{figure}[h!]
    \centering
    \vspace{-5pt}
    \includegraphics[width=1.0\linewidth]{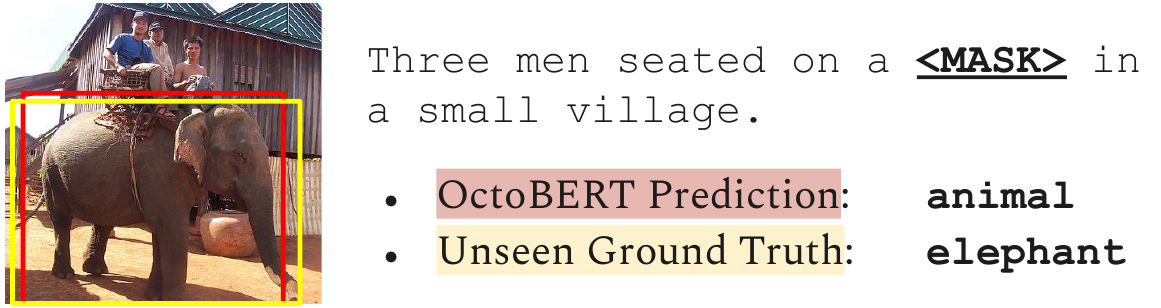}
    \vspace{-15pt}
    \caption{Although the word ``\texttt{elephant}'' is unseen to {\modelName}, the model is still able to localize the object in the image referred to by the \texttt{MASK}. \vspace{-10pt}}
    \label{fig::grounding}
\end{figure}

One important finding of the pre-trained model is the surprising performance in localizing the unseen words behind the \texttt{MASK}s.
As shown in Table~\ref{tab:result}, {\modelName} achieves a high Any-IoU of 56.3\% and Any-localization accuracy of 61.3\% for the unseen words, which are very close to its performance on the seen set and surpass baselines that have seen these words.
Moreover, as anticipated, since these words are held out during pre-training, {\modelName} fails to correctly unmask these unseen words, leading to a high log perplexity of 11.01 and low HR of 4.2, compared to that of 1.26 and 66.9 on the seen words.
Figure~\ref{fig::grounding} shows an example of such word-agnostic grounding.

This performance disparity in language modeling and referent localization on unseen words suggests that {\modelName} has developed a certain level of word-agnostic grounding, \textit{i.e.}, to locate the most likely referent of a word through both the linguistic context and the visual context, even if the word itself is never seen during pre-training.
A similar situation is faced by human language learners when inferring the grounded meaning of a new word, as we described earlier in Figure~\ref{fig::mapping}.
Our experiment demonstrates that, through grounded pre-training, it is possible for a vision-language system to acquire word-agnostic grounding ability, which opens up the opportunity to enable human-like fast mapping when learning new words.

\subsection{Few-Shot New Words Acquisition}

In this section, we task the model to acquire unseen words from a few samples of raw image-text pairs, without any bounding boxes or word-object mappings annotation.
As we have demonstrated the model's word-agnostic grounding, we seek to explore if this ability can be transferred to facilitate learning unseen words when a large amount of data and grounded supervision are no longer available.
Specifically, we perform few-shot learning on the pre-trained {\modelName} with only masked language modeling as the learning objective.
More hyper-parameter details are available in Appendix~\ref{appendix:fewshot}.

\paragraph{Learning New Words through Incremental Learning.}

\begin{figure}[h!]
    \centering
    \includegraphics[width=0.9\linewidth]{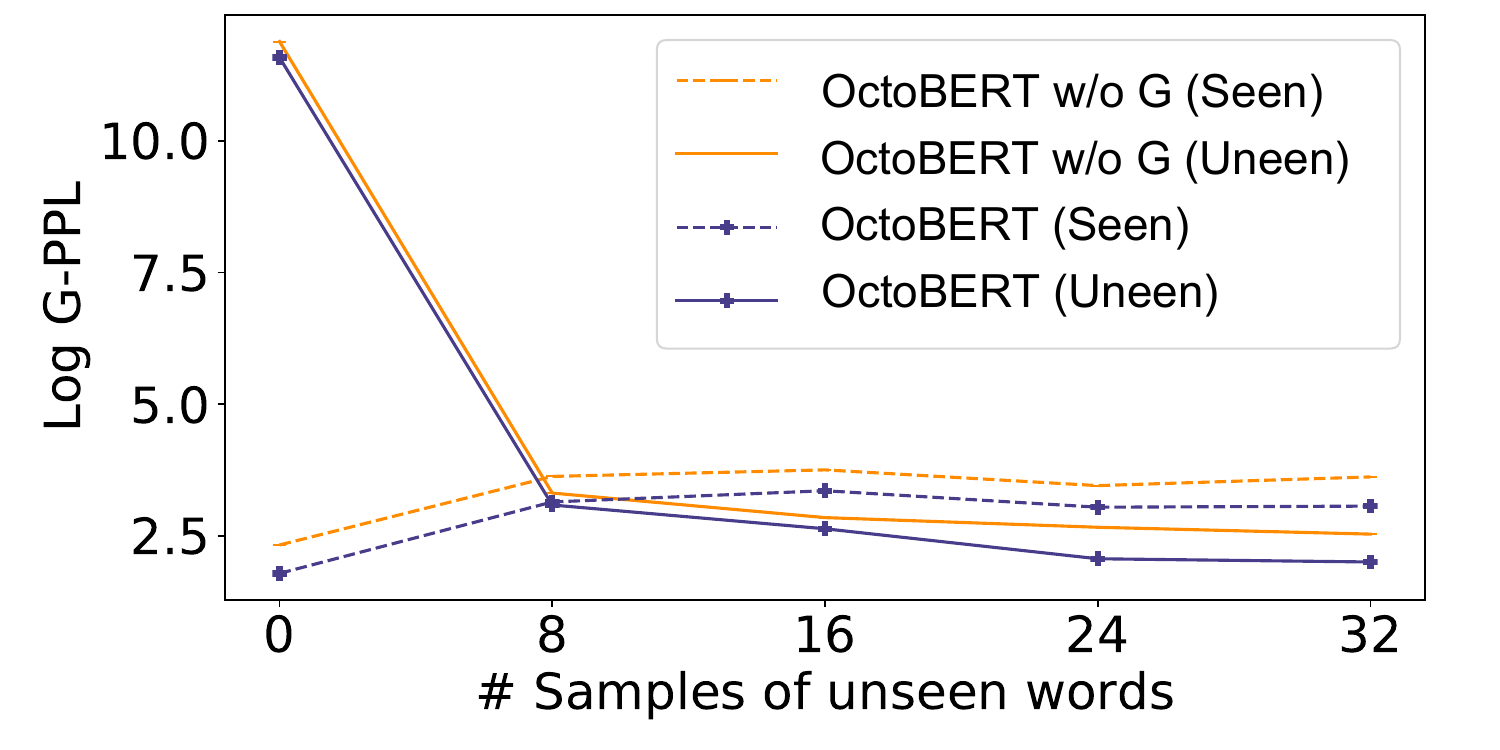}
    \vspace{-8pt}
    \caption{The log G-PPL (All-Protocol) of seen and unseen words in multi-class incremental learning, each unseen word with a sample size ranging from 8 to 32. \vspace{-8pt}}
    \label{fig::multiclass}
\end{figure}

We first explore the multi-class incremental learning setting, in which the pre-trained model is tasked to acquire the 31 unseen words from a few-shot learning session.
The experiment is repeated with sample sizes of 8, 16, 24, and 32 immediately after pre-training. 
As shown in Figure~\ref{fig::multiclass}, even with as few as 8 samples per word, {\modelName} can significantly bring down the grounded perplexity of unseen words, while mostly maintaining the grounded perplexity of the seen words without catastrophic forgetting. 
Compared to {\modelName} without the grounding objective, the full {\modelName} demonstrates better acquisition performance for unseen words. 
It's important to note that these few shot examples are text/image pairs without explicit grounding annotation. 
{\modelName} is able to quickly acquire grounded meanings of the new words (\textit{e.g.}, only with 8 examples) with a performance close to that of seen words. 

We further perform a word-specific controlled study with a one-class incremental learning setting.
We present results on two unseen words (\texttt{pizza} and \texttt{circular}) in Table~\ref{tab::oneclass}.
The complete results are available in Appendix~\ref{app::results}.

\begin{table}[!htp]
\centering
\scalebox{0.77}{
\begin{tabular}{ccccc}
\toprule
\multirow{2}{*}{\# Samp.} & \multicolumn{2}{c}{log G-PPL (\texttt{pizza})} & \multicolumn{2}{c}{log G-PPL (\texttt{circular})}   \\
\cmidrule(lr){2-3} \cmidrule(lr){4-5}
                            & {\small\modelName}           & {\small\modelName}$_\textrm{w/o G}$      & {\small\modelName}             & {\small\modelName}$_\textrm{w/o G}$       \\
\cmidrule(lr){1-1} \cmidrule(lr){2-2} \cmidrule(lr){3-3} \cmidrule(lr){4-4}  \cmidrule(lr){5-5}
0                           & 10.70               & 9.59            & 15.21                & 15.12            \\
8                           & \textbf{1.47}      & 2.21            & \textbf{1.59}        & 2.25             \\
16                          & \textbf{1.07}      & 2.54            & \textbf{1.07}        & 2.25             \\
24                          & \textbf{1.19}      & 1.25            & \textbf{1.55}        & 1.81             \\
32                          & \textbf{0.90}      & 1.18            & \textbf{1.23}        & 1.61            \\
\bottomrule
\end{tabular}
}
\vspace{-8pt}
\caption{The log G-PPL (All-Protocol) of unseen words in one-class incremental learning, each unseen word with a sample size ranging from 8 to 32. \vspace{-15pt}}
\label{tab::oneclass}
\end{table}

\subsection{Predictors of Model Behaviors}
\label{sec:analysis}

\begin{figure*}[!htp]
    \centering
    \begin{subfigure}[t]{.27\textwidth}
        \centering
        \includegraphics[width=1.06\linewidth]{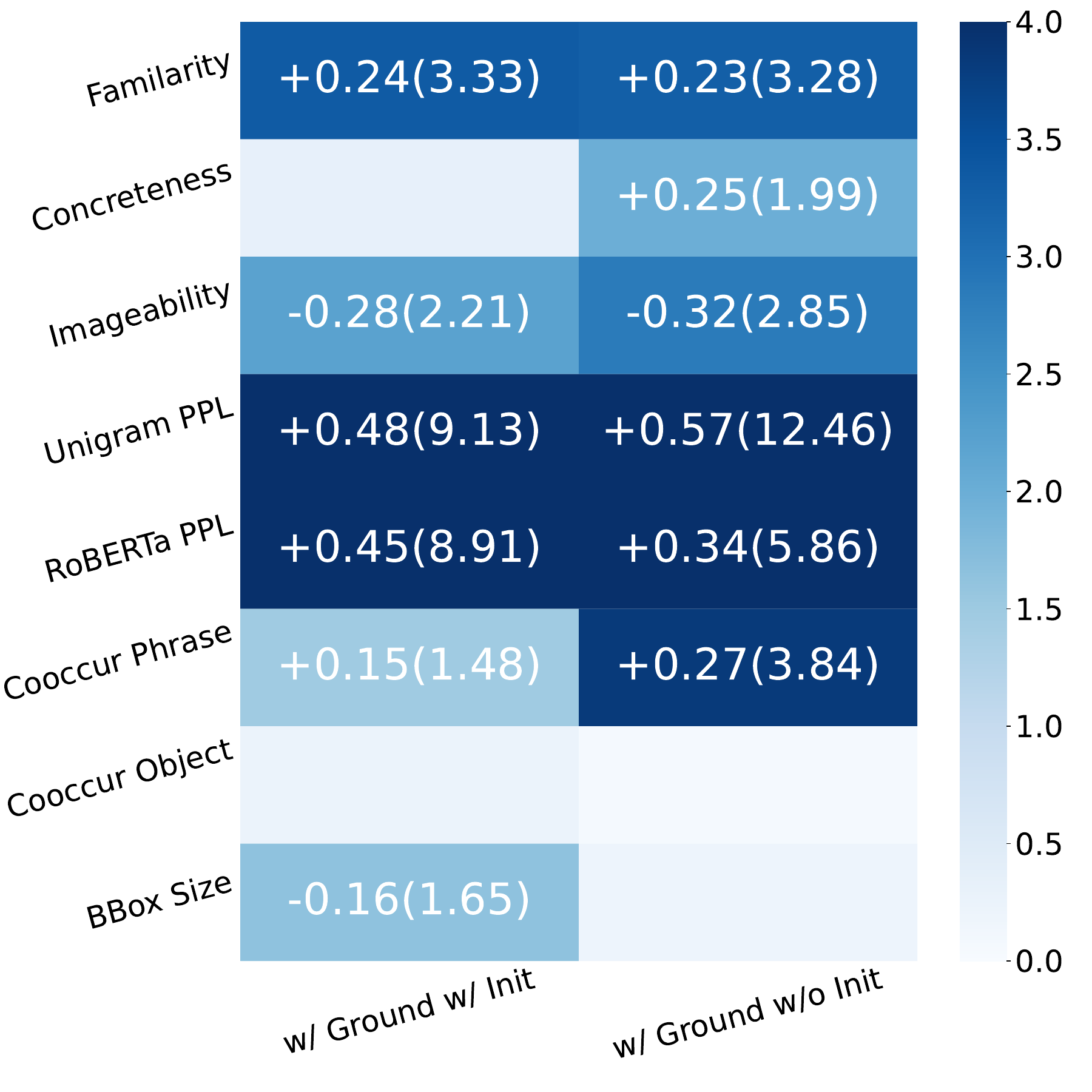}
        \vspace*{-0.5cm}
        \caption{$- \log p$-values for log G-PPL.}
        \label{fig::log-gppl}
    \end{subfigure}
    ~
    \begin{subfigure}[t]{.405\textwidth}
        \centering
        \includegraphics[width=1.06\linewidth]{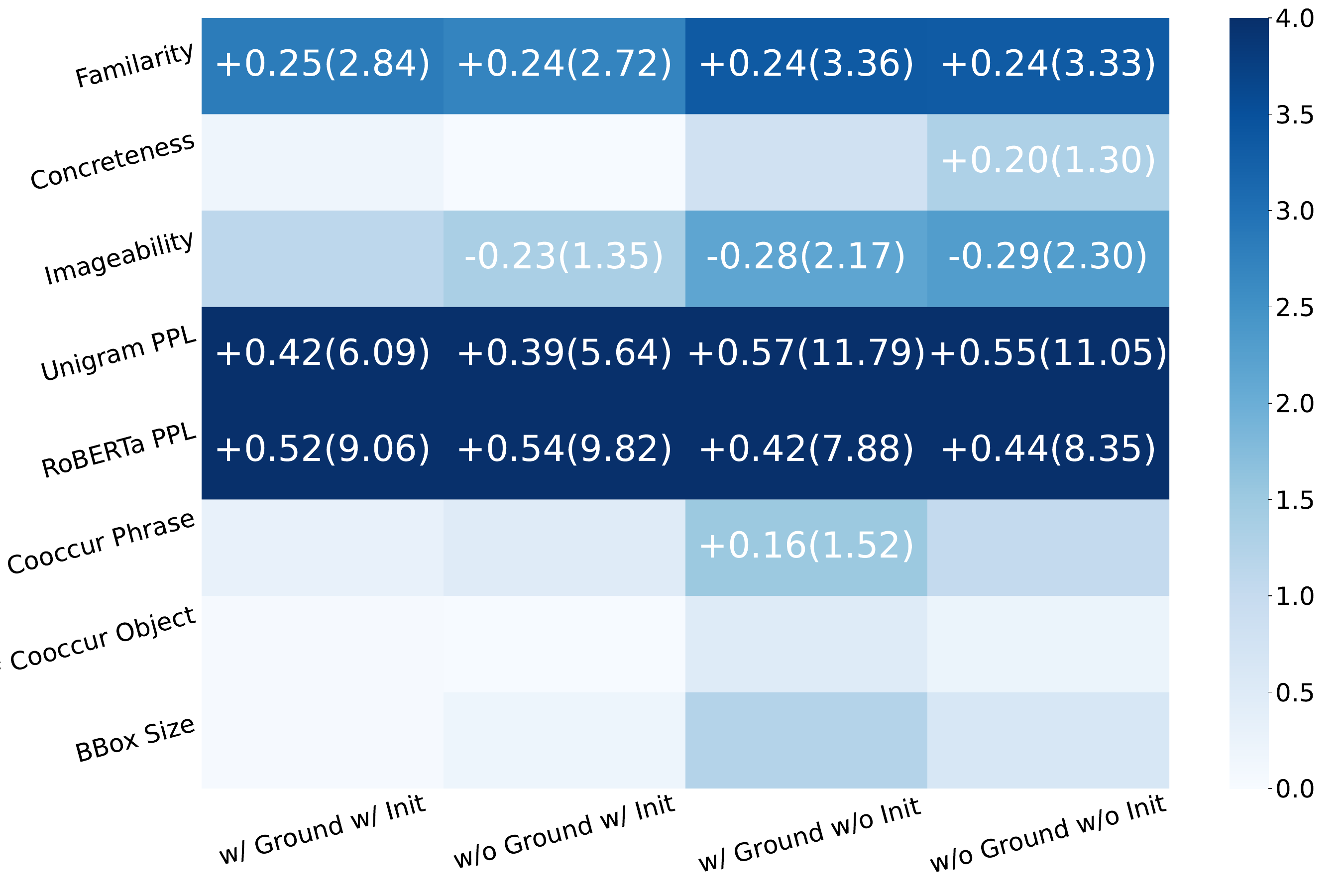}
        \vspace*{-0.5cm}
        \caption{$- \log p$-values for log PPL.}
        \label{fig::log-ppl}
    \end{subfigure}
    ~
    \begin{subfigure}[t]{.27\textwidth}
        \centering
        \includegraphics[width=1.06\linewidth]{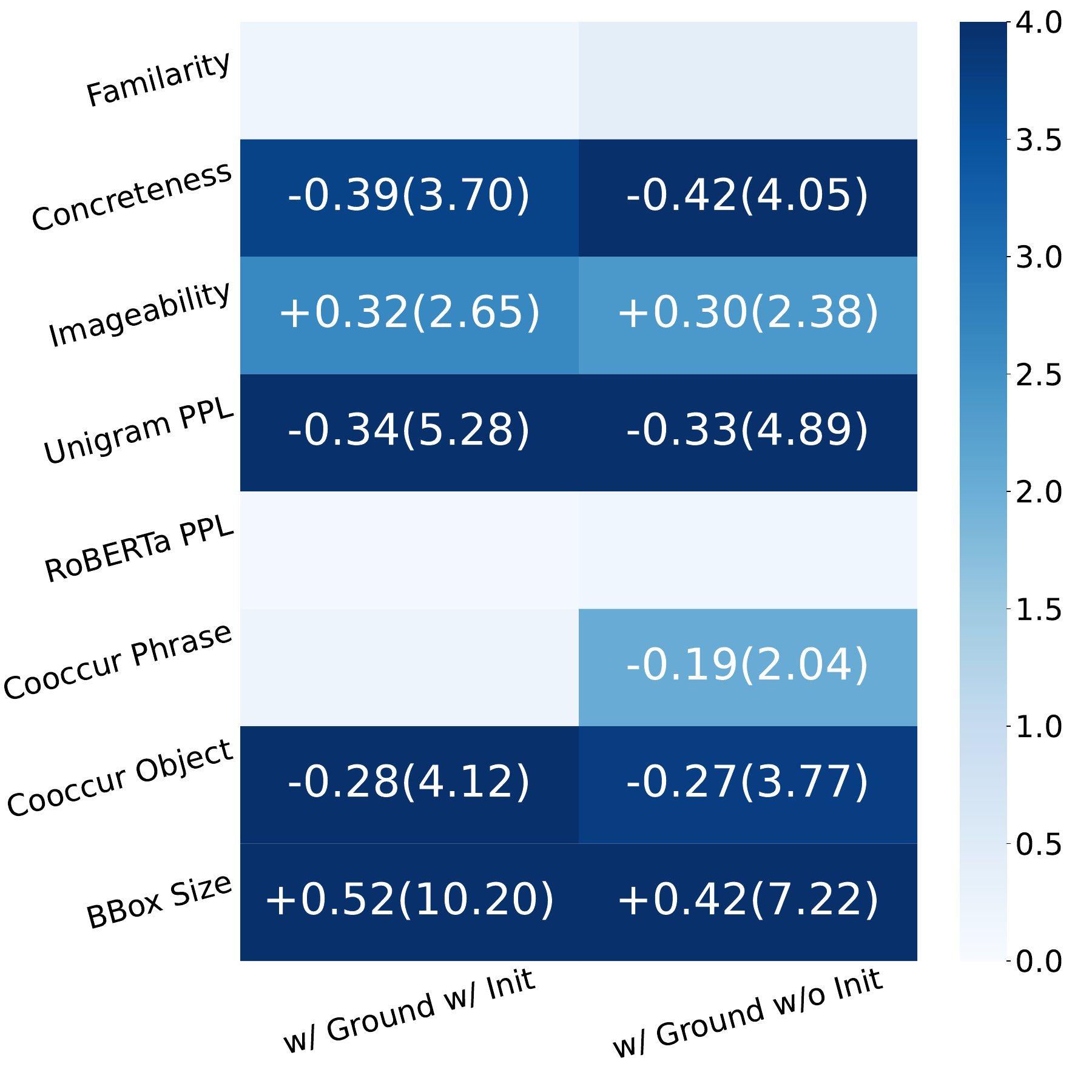}
        \vspace*{-0.5cm}
        \caption{$- \log p$-values for Any IoU.}
        \label{fig::any-iou}
    \end{subfigure}
    \vspace{-8pt}
    \caption{Heatmaps for statistical significance for each predictor towards each metric. The beta weights and signs are presented outside of the parentheses, and the negative log $p$-values are presented in the parentheses. Insignificant tests with $p > 0.05$, \textit{i.e.}, $-\log(p) < 1.30$, are discarded. w/(o) Init refers to the text encoder initialization. \vspace{-14pt}}
    \label{fig::behaviour-predict}    
\end{figure*}

There has been an interest to identify predictors that can explain/anticipate the performance or behavior of pre-trained language models~\citep{chang2022word}. 
This exploration not only offers valuable insights for future model development, but also serves as a cognitive inquiry to evaluate the extent to which language models align with human language acquisition patterns. 
In this section, we present the first work of this nature on vision-language models.
Specifically, we note that the \texttt{\modelName} model relies on a RoBERTa encoder, which might have already been equipped with prior linguistic knowledge.
To assess the cognitive alignment of vision-language models to human language acquisition, we additionally pre-trained the \texttt{\modelName} and \texttt{\modelName}$_\textrm{w/o G}$ models with a randomly initialized RoBERTa encoder.

To comprehensively capture various aspects of words, we carefully select eight distinct predictors that encompass intrinsic psycho-linguistic characteristics, distribution patterns within the training corpus, and visual representations within the training images.
We select 3 \textbf{psycho-linguistic predictors}, each collected and normalized from the MRC Database~\citep{coltheart1981mrc}: 
\vspace{-0.2cm}
\begin{itemize}[leftmargin=*]
    \setlength\itemsep{-0.4em}
    \item \texttt{Familiarity}, the degree of familiarity or exposure people have to words; 
    \item \texttt{Concreteness}, the degree to which words have a perceptible physical referent or are associated with tangible objects or experiences; 
    \item \texttt{Imageability}, the degree to which words elicit people's mental imagery.
\end{itemize}
\vspace{-0.15cm}
Another 3 \textbf{linguistic predictors} are considered: 
\vspace{-0.2cm}
\begin{itemize}[leftmargin=*]
    \setlength\itemsep{-0.25em}
    \item \texttt{Unigram perplexity}; 
    \item \texttt{RoBERTa perplexity}, where RoBERTa is fine-tuned on the captions to serve as the upper bound of unimodal language model performance;
    \item \texttt{\# Co-occur phrases}, the average number of co-occurring groundable phrases in a caption.
\end{itemize}
\vspace{-0.15cm}
We finally choose 2 \textbf{perceptual predictors}:
\vspace{-0.2cm}
\begin{itemize}[leftmargin=*]
    \setlength\itemsep{-0.25em}
    \item \texttt{\# Co-occur objects}, the average number of co-occurring objects in an image;
    \item \texttt{Bbox size}, the average proportion of an image occupied by the bounding boxes of the referents.
\end{itemize}
\vspace{-0.15cm}
To assess the statistical significance of each predictor, we performed linear regressions with likelihood ratio tests on different variants of models.
Similar to~\citet{chang2022word}, we compare the overall regression including the target predictor to a regression that included all predictors except the target. 
We additionally present the beta weights (with signs) to capture the magnitude and direction of the correlation.
Figure~\ref{fig::behaviour-predict} displays heatmaps indicating the statistical significance (in terms of negative logarithmic $p$-values) of each predictor concerning Log G-PPL, Log PPL, and Any IoU.
Insignificant tests are omitted from the figure.

\vspace{-0.1cm}
\paragraph{Correlation with Linguistic and Perceptual Predictors.}
Our findings revealed a positive correlation between the unigram and RoBERTa log perplexity and the models' log perplexity, both for grounded and ungrounded scenarios. 
This indicates that vision-language models still heavily rely on distributional statistics, similar to unimodal models.
While the ungrounded perplexity showed little correlation with perceptual predictors, the Any IoU demonstrated a significant correlation with the number of co-occurring objects and average sizes of bounding boxes. 
This suggests concepts that are visually salient and less perceptually ambiguous are easier to localize and acquire, consistent with human learners~\cite{smith2008infants}.

\vspace{-0.1cm}
\paragraph{Correlation with Psycho-linguistic Predictors.}
Counter-intuitively, there was a positive alignment between the human perceived familiarity of words and the machine's perplexities, \textit{i.e.}, the more familiar humans are with a word, the more perplexed models get.
This contrasts with the ideal cognitive plausibility of language acquisition in humans. 
This discrepancy implies that current vision-language models may not fully achieve cognitive plausibility, which might be explained by the fact that many concepts (\textit{e.g.}, wild animals, musical instruments) appear abundantly in internet images but not in daily lives.
In terms of imageability, it aligned well with human intuition, exhibiting a positive correlation with Any IoU and a negative correlation with perplexities. 
However, the concreteness predictor surprisingly exhibited the opposite correlation. 
This discrepancy could be attributed to the nuanced distinction between imageability and concreteness. 
For instance, while ``\texttt{hat}'' is concrete because it refers to a tangible object, it also possesses visual diversity due to its generality (\textit{e.g.}, many types of hats which look very differently), making it challenging to acquire. 
Conversely, ``\texttt{blue}'' is more imageable as it easily evokes a color, relatively stable, despite not referring to a specific tangible object.
To learn the meaning of ``\texttt{hat},'' a human language learner may benefit from physically interacting with the object, and understand that the hat is an item to cover for the head, regardless of its visual appearance. 
To address this gap, a potential future direction could involve developing language learning agents that acquire words through physical interactions rather than passive perception, allowing for a more comprehensive understanding of word meanings.

\vspace{-0.1cm}
\section{Related Work}
\label{sec:background}

\vspace{-0.1cm}
\paragraph{Grounding Language to Vision}

Referential grounding plays a central role in classic lexicon acquisition problem~\citep{gleitman1994acquisition,clark1995lexicon}.
Primarily, researchers focused on grounding words to their meaning symbols, building learning mechanisms using specific mental biases to simulate children's word acquisition, and giving computational accounts for psycholinguistic phenomena~\citep{siskind1996computational,regier2005emergence,goodman2007bayesian,fazly2010probabilistic}. 
Early research in this area focused on visual grounding through two main approaches: first, by developing statistical and neural mappings from object categories to lexical items~\citep{roy2002learning,yu2005emergence,xu2007word,yu2007unified,yu2013grounded}, and later by incorporating more sophisticated visual features to establish connections with lexicon, syntax, and complex semantic structures~\citep{qu2010context,shi2019visually,jiayuan2019neuro,mao2021grammarbased,pratt2020grounded}.
These studies are usually in a closed world with limited vocabulary~\citep{krahmer2019computational}, and words are usually isolated from the natural context of use.
More recently, there exist multi-modal understanding tasks, \textit{e.g.}, object retrieval~\citep{guadarrama2014open,hu2016natural}, referring expression comprehension~\citep{liu-chai-2014,yu2016modeling,mao2016generation,wu2020phrasecut}, and phrase grounding~\citep{plummer2015flickr30k} to map text to corresponding objects.
Our setup relates to this line as we position referential grounding as an explicit word-referent mapping problem. 
We focus on efficient learning of grounded lexical semantics through fast mapping in modern vision-language models, a more scalable but realistic challenge faced by AI agents.

\vspace{-0.1cm}
\paragraph{Vision-Language Pre-training}

Distributional word representations can be acquired through language modeling, and developing language models from visual data has been extensively studied by the community~\citep{chrupala2015learning,lazaridou2015combining,li2017learning,suris2020learning}.
Recent years have seen increasing research to enrich language representations with visually-augmented language modeling~\citep{tan2020vokenization,lu2022imagination,wang2022visually} and to learn multimodal representations with vision-language pre-training (VLP)~\citep{du2022survey}.
We are particularly interested in VLP models with fine-grained grounding objectives, \textit{e.g.}, Word-Region Alignment (WRA).
These models either pre-train with weakly supervised alignment algorithms like optimal transport that matches words with patches~\citep{kim2021vilt} or proposals from a frozen detector~\citep{chen2020uniter,su2020vlbert}, or perform explicit word grounding by pre-training a language-conditioned detector~\citep{kamath2021mdetr,li2022grounded,zhong2022regionclip,dou2022coarse}.
Our model falls along this line, which jointly performs language modeling, object localization, and grounding during pre-training, rather than relying upon a pre-existing object detector. 

\vspace{-0.1cm}
\paragraph{Vision-Language Tasks}

To evaluate vision-language systems, many downstream tasks have been formulated. 
Some related formulations are summarized in Table~\ref{tab:tasks} in the Appendix.
While demonstrating some vision-language capabilities, these downstream tasks provide limited insights into whether these models truly capture the grounded meaning of words with respect to the external environment.
Our task design specifically targets the machine’s ability to predict words and ground words to perception.
More akin to our formulation is the vision-based language modeling task~\citep{jin2020visually} in a continual learning setting.
Our work differs mainly in two aspects.
First, the proposed task only predicts masked tokens based on the visual context, which leaves the referential uncertainty (\textit{i.e.}, grounding) unattended (\textit{e.g.}, in Figure~\ref{fig::grounding-task}, correct prediction of the word ``\texttt{boat}'' does not guarantee correct grounding).
Also, \citet{jin2020visually} focuses on compositionality, while we seek to address few-shot grounded word learning when unseen words are encountered after pre-training.

\paragraph{Open-Vocabulary Object Detection}

Early works formulate fast mapping of new words as a zero-shot object classification problem, which aims to generalize from known object labels to unknown ones~\citep{socher2013zero,frome2013devise,elhoseiny2013write,lazaridou2014wampimuk}.
The setting later extends to a localization task, referred to as zero-shot object detection (ZSD)~\citep{bansal2018zero,zhu2019zero,zhu2020don,rahman2020improved}.
More recently, open-vocabulary object detection (OVD)~\citep{zareian2021open,gu2022openvocabulary,du2022learning,minderer2022simple} combines ZSD with weakly supervised object detection (WSD) to address the unrealistic constrain of traditional zero-shot settings.
OVD assumes the availability of coarse-grained image-caption pairs, and attempts to generalize from limited fine-grained annotation of object categories to unseen ones.
Nevertheless, this line of work positions words as object categories and isolates them from their linguistic context (\textit{e.g.}, sentences).
Our setup instead challenges models to perform language modeling in human-generated captions.
\section{Conclusion and Future Work}
\label{sec:conclusion}

The connection between language and their referents captures the grounded meaning of words, and an explicit treatment is key to empowering efficient open-world language learning abilities in humans and AI agents. 
This work introduces Grounded Open Vocabulary Acquisition (\texttt{GOVA}), a scalable formulation to examine grounding and fast mapping in open-world grounded language learning.
We propose {\modelName}, a novel visually grounded language model to investigate a paradigm where the model initially acquires grounding ability during pre-training and subsequently applies this ability to quickly learn new words without explicit grounding supervision.
Our empirical findings highlight the significance of visual grounding in neural word acquisition. Especially, we find that pre-trained {\modelName} can serve as a foundation for fast mapping of novel grounded words via few-shot learning.
We also conduct a comprehensive analysis to explore potential predictors influencing the performance of vision-language models, revealing both consistent and surprising behaviors with respect to human language learning patterns. 
These insights pave the way for future research in grounded language learning in the open world.

\section*{Limitations}

In this work, we limit ourselves to object-centric grounding, which ignored that language can ground events, attributes, manners, mental states, etc.
The grounded meaning of some groundable words, especially \texttt{ADV}s, \texttt{NUM}s, \texttt{VERB}s, and \texttt{PRON}s, cannot be fully captured by the bounding boxes alone.
Future work should explore better task formulations to study the acquisition of their grounded meanings.
An exciting future work along this line is to extend the setting from images to videos and physical interactions with the environment, and to incorporate the rich temporal dynamics of the world for language acquisition.
In addition, we ignored the social aspects of language learning, where children infer the referents of words from their caregivers through communication~\citep{carpenter1998social,bloom2000children}.
Future work could also investigate grounded word acquisition from natural dialogue.

\vspace{-0.08cm}
\section*{Ethics Statement}

This project does not involve any research artifacts generated through human subject studies. 
Despite the considerable promise of {\modelName}, it is crucial to examine its ethical and societal implications. 
The computational model relies on pre-trained language models and extensive text-image datasets, which could contain hidden biases that may result in fairness problems within the algorithms. 
By recognizing and actively addressing these implications, we aim to increase awareness among practitioners if the model is deployed as a language-learning agent in the future.

\vspace{-0.08cm}
\section*{Acknowledgments}

This work was supported in part by NSF IIS-1949634, NSF SES-2128623, and by the Automotive
Research Center (ARC) at the University of Michigan. 
The authors would like to thank the anonymous reviewers for their valuable feedback.

\bibliography{reference}
\bibliographystyle{acl_natbib}

\clearpage
\appendix

\section{\texttt{GOVA} Dataset Details}
\label{appendix:data}

\begin{figure*}[t]
    \centering
    \includegraphics[width=1.0\linewidth]{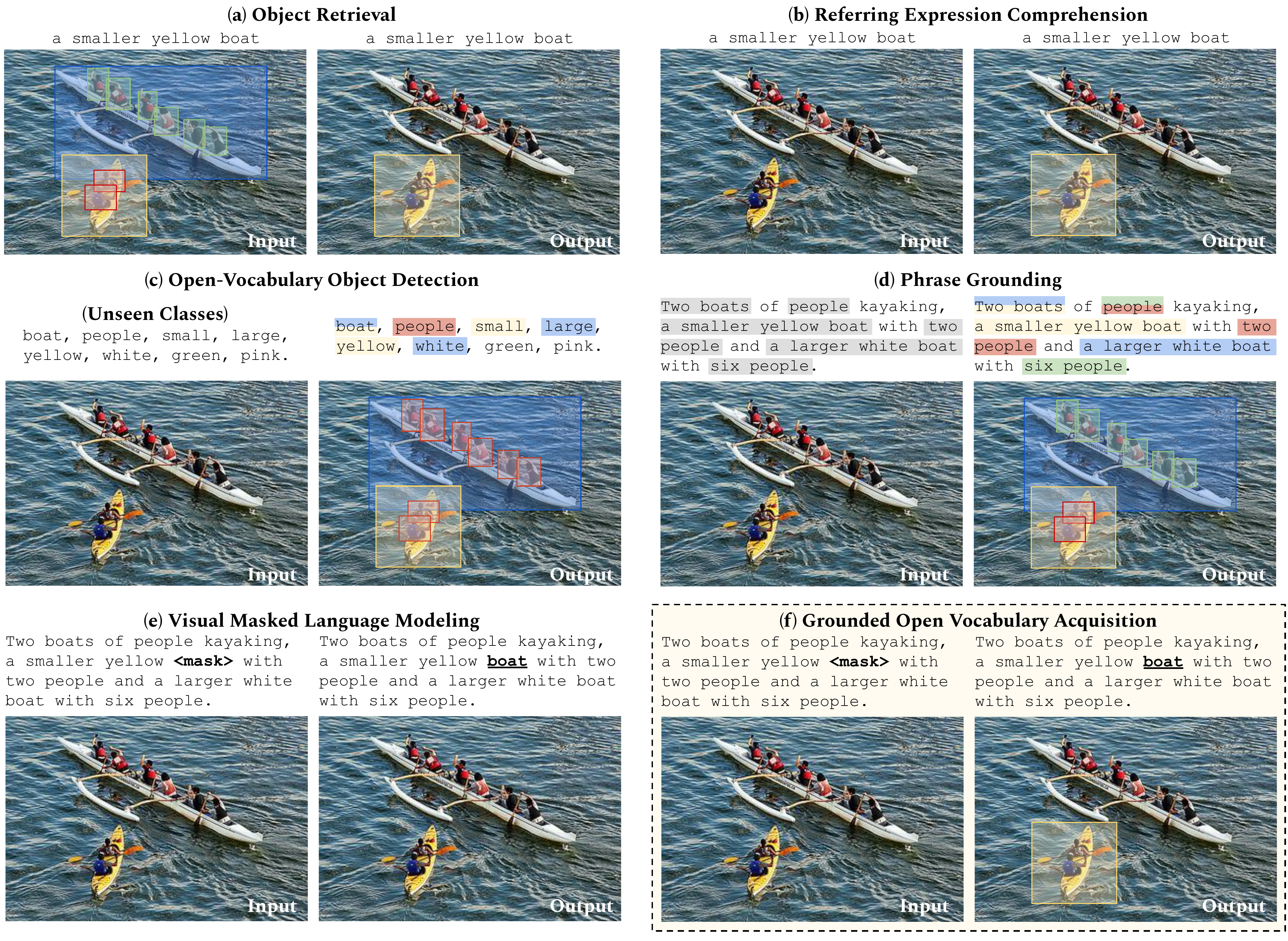}
    \vspace*{-18pt}
    \caption{An illustrated comparison of task formulations related to grounded language learning.}
    \label{appendix::task}
\end{figure*}

\subsection{Illustrated Comparison of Setting}

\begin{table*}[!htp]
\scalebox{0.52}{
\begin{tabular}{cccccc}
\toprule
\textbf{Tasks (Inference Time)}           & \textbf{Language Input}                                & \textbf{Visual Input}                         & \textbf{Language Output} & \textbf{Vision Output}                 & \textbf{Example Dataset(s)}                                        \\ \midrule
Masked Language Modeling             & Cloze Test                                             & -                                             & Missing Word             & -                                      & WikiText-103~\citep{merity2017pointer}                                                    \\
Knowledge Probing             & Cloze Test                                             & -                                             & Missing Word             & -                                      & LAMA~\citep{petroni2019language}                                                    \\
Reading Comprehension                & Context, Cloze Test                                         & -                                             & Missing Word             & -                                      & LAMBADA~\citep{paperno2016lambada} \\ 
\cmidrule(lr){1-1} \cmidrule(lr){2-6}
Image Captioning                     & -                                                      & Image                 & Caption                  & -                                      & Flickr30k~\citep{young2014image} \\
Fill-in-the-Blank VQA                & Cloze Test, (Choices)                                    & Image/Video                                   & Missing Text (Choice)    & -                                      &  FIBER~\citep{castro2022fiber}                                                   \\
Visual Masked Language Modeling & Cloze Test                                             & Image, Bounding Boxes & Missing Word             & -                                      & VisCOLL~\citep{jin2020visually}                                                 \\
\cmidrule(lr){1-1} \cmidrule(lr){2-6}
Object Retrieval                     & Referring Expression           & Image, Bounding Boxes                         & -                        & Bounding Boxes &   ReferIt~\citep{kazemzadeh2014referitgame}, \\
Referring Expression Comprehension   & Referring Expression           & Image                                         & -                        & Bounding Boxes &  RefCOCO*~\citep{yu2016modeling,mao2016generation}                      \\ \cdashlinelr{6-6}
Phrase Grounding                     & Caption, Referring Expressions & Image                                         & -                        & Bounding Boxes & Flickr30K Entities~\citep{plummer2015flickr30k}                                      \\
\cmidrule(lr){1-1} \cmidrule(lr){2-6}
Object Detection                     & Seen Classes                                                      & Image   & Classes                  & Bounding Boxes &                                                         \\
Zero-Shot Object Detection           & Unseen Classes                                                      & Image & Classes                  & Bounding Boxes &                                                         \\
Open-Vocabulary Object Detection     & Pre-training Vocabulary                                & Image                 & Words                    & Bounding Boxes & \multirow{-3}{*}{ LVIS~\citep{gupta2019lvis}}                      \\
\cmidrule(lr){1-1} \cmidrule(lr){2-6}
Grounded Open Vocabulary Acquisition    & Cloze Test                                             & Image                                         & Missing Word             & Bounding Boxes & \textbf{\texttt{GOVA} (Ours)}                                             \\ \bottomrule
\end{tabular}
}
\caption{Comparison of task formulations related to grounded language learning. \vspace{-10pt}}
\label{tab:tasks}
\end{table*}

We present an illustrated comparison of task formulations related to language grounding and grounded language learning in Figure~\ref{appendix::task}.
Among these task formulations, our Grounded Open Vocabulary Acquisition (\texttt{GOVA}) task is the only one that challenges vision-language systems to perform visually grounded and object-centric language modeling.
The formulation is natural and simple, with fundamental requirements on computational models to perform masked language modeling and object localization, and thus is particularly good for zero-shot analysis.

\subsection{Evaluation Protocols Explained}
\label{appendix:iou}

We present an adequate evaluation protocol for {\em grounded} word acquisition in the main paper. 
This section provides more in-depth explanation for the metrics and implementation details for reproducibility purposes.

\paragraph{Perplexity Metric Details}

We follow prior practice in cloze tests~\citep{salazar2020masked,jin2020visually} to evaluate the perplexity of a word $w$.
We use log pseudo-perplexity in masked language modeling, defined as
\begin{equation*}
    \log\text{PPL}(w) = -\log P(w|x_{\text{img}},x_{\text{cap}})
\end{equation*} 
However, the majority of the language models employ sub-word tokenization methods to segment and encode text.
In particular, one lexical word can be segmented into several tokens, and different tokenizers can lead to different tokens for the same input.
We thus introduce a tokenizer-dependent measure for perplexity.
For tokenizer $T$, we represent the $N$ tokens of word $w$ as $T(w)$ and
\begin{equation*}
    \log\text{PPL}(w) = -\frac{1}{N}\sum_{t\in T(w)}\log P(t|x_{\text{img}},x_{\text{cap}})
\end{equation*}

\paragraph{IoU Metric Details}

we face the same challenge as~\citet{kamath2021mdetr} where multiple referents are possible for a masked word.
In a similar manner, we adopt the Any-Protocol and All-Protocol to evaluate the grounded detection task.
Assuming $n$ ground truth bounding boxes $B = \{b_1, b_2, \cdots, b_n\}$ and $m$ predicted bounding boxes $\widetilde{B} = \{\widetilde{b_1}, \widetilde{b_2}, \cdots, \widetilde{b_m}\}$.
The intersection-over-union (IoU) under Any-Protocols is defined as the average IoU of the best matching predicted bounding box for each ground truth object:
\begin{equation*}
    \text{IoU}_{\text{any}} = \frac{1}{n} \sum_{i\in\{1,2,\cdots,n\}} \max_{j\in\{1,2,\cdots,m\}} \text{IoU}(b_i, \widetilde{b_j})
\end{equation*}

The intersection-over-union (IoU) under All-Protocols is defined as the IoU between the joint bounding box of ground truth and predicted bounding boxes:
\begin{equation*}
    \text{IoU}_{\text{all}} = \text{IoU}(\cup B, \cup \widetilde{B})
\end{equation*}

\subsection{Word List}
\begin{itemize}
    \item 60 words are in the seen-set, each with 80 test cases: baby, ball, beach, bench, bike, black, blond, blue, boy, brown, building, car, child, dark, dog, dress, face, female, field, floor, food, girl, glasses, grass, gray, green, guitar, guy, hair, hand, hat, head, horse, jacket, jeans, lady, large, little, long, man, orange, pants, person, player, red, shirt, sidewalk, sign, small, snow, street, striped, table, top, wall, water, white, woman, yellow, young.

    \item 31 words are in the unseen-set, each with 50 test cases\footnote{a few words (product, steep, telephone) has one less test case due to the availability of Flickr30K Entities Dataset.}: aged, bamboo, barefoot, brush, button, cafe, cheese, circular, classroom, crosswalk, diverse, doctor, donkey, elephant, fluffy, foreign, gym, heart, newborn, pan, pizza, product, security, sink, star, steep, stove, student, teacher, telephone, warm.
\end{itemize}

\section{Computational Model Details}
\label{appendix:model}

\subsection{Pre-training Objectives}
\label{appendix:loss}

\paragraph{Masked Language Modeling (MLM).} 

The MLM head can be placed at multiple possible places, and our design is an exploration after preliminary experiments on smaller-scale training.
We strictly follow the setup of RoBERTa to implement the MLM head with a two-layer MLP, based on the implementation of \texttt{huggingface}\footnote{\url{https://huggingface.co/docs/transformers/model_doc/roberta}}.
Words in groundable phrases are masked with a probability of 0.4 and those in non-groundable regions are masked with a lower probability of 0.1.
For a token selected to mask, we follow RoBERTa to assign a probability of 80\% to replace with \texttt{MASK}, 10\% with a random token, and 10\% to do nothing.

\paragraph{Object Localization (OL).} 
We follow MDETR to decode object embeddings with a three-layer MLP to produce bounding boxes.
Similar to most prior work, we apply a filter over boxes with confidence below 0.7.
In our framework, this means that the object corresponds to the no-object label $\varnothing$ (Figure~\ref{fig::model}) with a probability over 0.3.
We strictly follow DETR to perform bipartite matching between proposed boxes and ground truth boxes with a Hungarian loss.
The predicted boxes are optimized towards ground truth by the generalized intersection-over-union (GIoU) loss and the L1 loss.

\paragraph{Grounding.} 
In positional alignment, the model learns to map each object representation to tokens in the sentence with a fixed length of 257, which could possibly be a \texttt{MASK} or an additional no-object label $\varnothing$ (Figure~\ref{fig::model}).
The object and the token are considered a match given a mapping probability over 0.1.
We use a fully-connected layer to predict the distribution over token positions with cross-entropy loss.
In semantic alignment, the model learns to bring word embeddings closer to the object embeddings that they ground to, and push the unrelated pairs farther. 
We strictly follow the contrastive loss function defined in MDETR for every object and groundable token for this purpose.

\subsection{Few-shot Learning Details.}
\label{appendix:fewshot}

Since no bounding box or word-object mappings annotation is available, we train {\modelName} with only masked language modeling (MLM) in few-sample new word learning.
We reduce the batch size to 8 considering the fewer number of samples, and set the convergence criteria to a fixed number, \textit{i.e.}, 50 steps.
All the rest of the experimental settings remain the same as pre-training.

\section{Experiment Reproducibility}
\label{appendix:experiment}

\subsection{{\modelName} Implementation Details}
Our {\modelName} model mainly consists of one cross-modal transformer with inputs from uni-modal encoders from image and text domain. Specially, we select the ResNet-50~\cite{he2016deep} pretrained on ImageNet from \texttt{TIMM}\footnote{\url{https://github.com/rwightman/pytorch-image-models}} as the image encoder, and RoBERTa-base \cite{liu2019roberta} from \texttt{huggingface}\footnote{\url{https://huggingface.co/docs/transformers/model_doc/roberta}} as the text encoder. The cross-modal encoder and two decoders each consists of 4 transformer blocks with 8 attention heads, an input and output dimensionality of 512, and an inner-layer dimensionality of 2,048. 
Besides, 50 learnable object queries are included to query the cross-modal decoder to generate bounding box proposals.

\subsection{Hyper-parameter Decisions}

\label{appendix:hyper}

We include the major hyper-parameter tuning decisions for reproducibility purpose.
For more details, please refer to the supplementary codes.
\begin{itemize}[leftmargin=*]
    \setlength\itemsep{-0.25em}
    \item Learning Rate:
    \begin{itemize}[leftmargin=*]
        \item Image Encoder: frozen
        \item Text Encoder: $1\times 10^{-5}$
        \item Multi-modal Transformer: $1\times 10^{-4}$
    \end{itemize}
    \item Batch Size: 128
    \item Pre-training Loss Coefficients:
    \begin{itemize}
        \item MLM Loss: 32
        \item Cross Entropy for Positional Alignment: 1
        \item Contrastive Loss for Semantic Alignment: 1
        \item L1 Localization Loss: 5
        \item GIoU Localization Loss: 2
    \end{itemize}
    \item Few-shot Learning:
        \begin{itemize}[leftmargin=*]
        \item Batch size: 8
        \item Other Hyper-parameters: Same as Pre-training
    \end{itemize}
\end{itemize}

\subsection{Computational Resources}

Our {\modelName} models is pre-trained on 8 NVidia A40 GPUs. 
With mixed-precision pre-training and a batch size of 128, {\modelName} was trained for 150,000 steps where each step takes about 1.4 second.

\subsection{Evaluation on \texttt{GOVA}}

\paragraph{{\modelName}}

For our proposed {\modelName} model, given a \texttt{GOVA} test, with its corresponding image and textual cloze pair passing into the model, the bounding box predictions are generated by keeping only the bounding box proposals that are mapped to at least one masked token within the cloze, while the masked token prediction results are directly decoded from its language modeling head.

\paragraph{VisualBERT}
For the “Detect-and-Recognize” baseline model VisualBERT, we use 
phrase-grounding fine-tuned version of VisualBERT to perform object localization, and, as it lacks the language modeling head, another vanilla pre-trained VisualBERT to perform mask token prediction. Specifically, for the bounding box localization part, we treat it as a standard phrase grounding task and follow~\cite{li2019visualbert} to select the top-1 bounding box prediction in the last masked token as the output.

\paragraph{ViLT+MDETR}
For the “Produce-and-Localize” baseline model ViLT + MDETR, in stage one, we feed the input image and text into ViLT, collecting its top-1 cloze token prediction result. Then, at stage two, the input image and ViLT-completed text are fed into MDETR, performing phrase-grounding to localize the object associated with the original cloze. Finally, the cloze token prediction result from ViLT together with the bounding box proposals from MDETR are used for \texttt{GOVA} evaluation.

\section{Addendum to Results}
\label{app::results}

\subsection{Ablation Study}

We performed an ablation study on several {\modelName} model variants to pinpoint what makes our {\modelName} model effective.
These included models without language encoder initialization (w/o Init), without grounding objective (w/o G), without any object-centric representation (w/o O), and a text-only setup without any vision input (w/o V).
For consistency, we control the number of transformer layers and the number of parameters for each variation.
Despite tweaking various hyperparameters, no significant improvements were observed. 
As a result, we retained the same hyperparameters as in the {\modelName} model.

\begin{itemize}[leftmargin=*]
    \setlength\itemsep{-0.25em}
    \item w/o G: This refers to the model variant without grounding loss, as has already been described in Section \ref{sec:baselines};
    \item w/o O: This variant excludes all object-centric representations, retaining only the masked language modeling (MLM) objective. With this model, the object decoder transformer is unnecessary, thus no grounding nor localization is performed. Instead, we consolidate all 12 transformer blocks into the multi-modal encoder and directly attach the MLM objective to it.
    \item Text-Only: This text-only model operates without any vision input or supervision, reducing it to a unimodal language model (RoBERTa) with 12 additional transformer blocks.
\end{itemize}

\begin{figure}[!ht]
    \centering
    \includegraphics[width=1.0\linewidth]{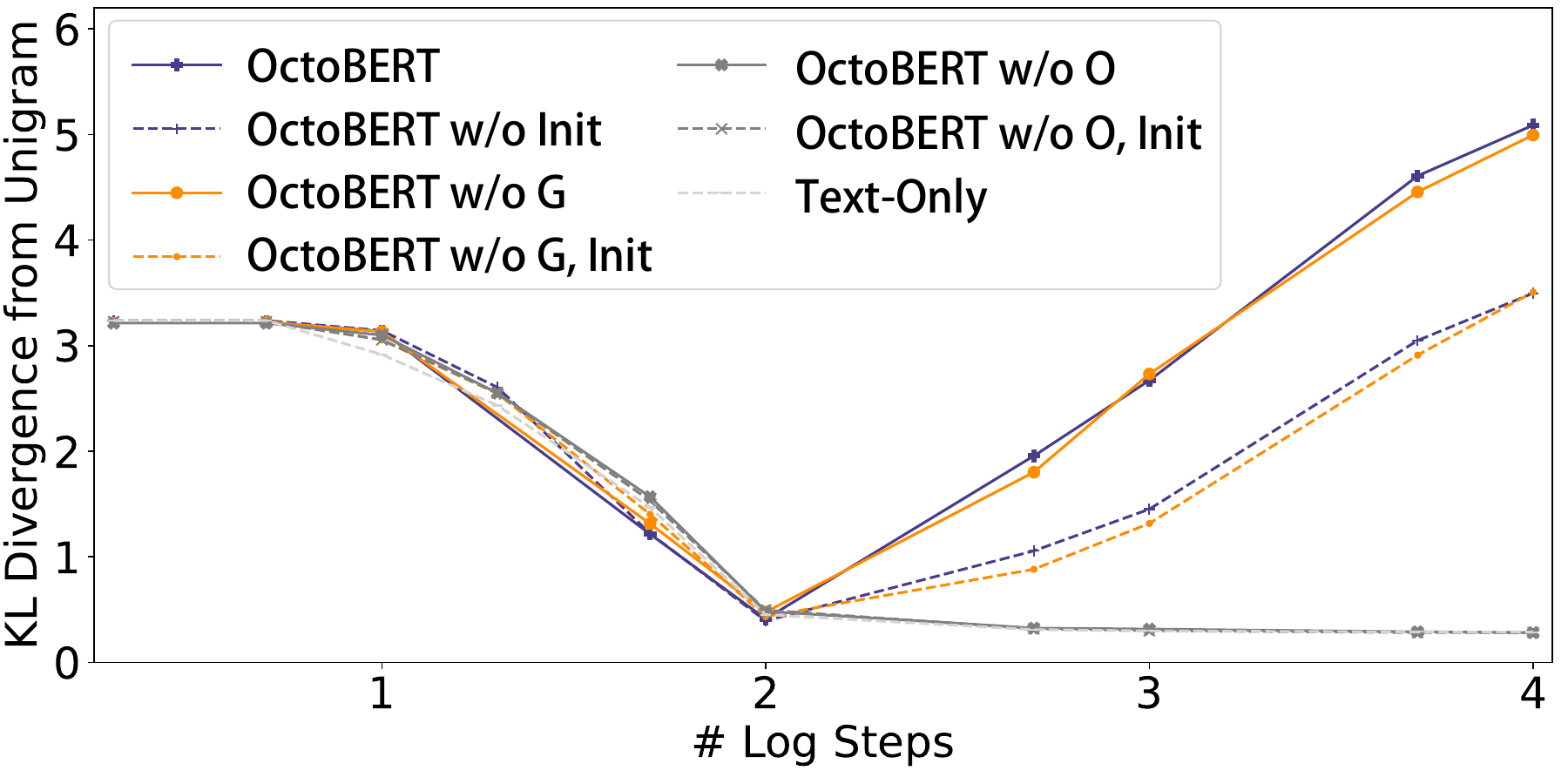}
    \caption{KL-divergence between model's token prediction and the unigram distribution of the training corpus.}
    \label{fig::kl}
\end{figure}

Following the analysis of~\citet{chang2022word} in unimodal language models, we present the KL-Divergence between the model predictions and the unigram distribution in Figure~\ref{fig::kl}.
An immediate observation is that all variants converge to the shallow unigram statistics at around $10^2$ steps of pre-training.
This aligns with the findings of~\citet{chang2022word} that unimodal language models would converge to unigram before acquiring more complicated contextual representations.
We noticed that in both text-only and {\modelName}$_\textrm{w/o O}$ cases where MLM is the only pre-training objective, the models tend to stay around the unigram word distribution even with $10^4$ steps of training. 
However, variants with an object-centric representation quickly departed from the unigram distribution.
Comparatively, models with language model initialization move quickly away from the unigram distribution, and models with a grounded objective have a marginally faster deviation.
These results confirm that vision-language models can benefit from unimodal pre-training on a large corpus, and that performing language modeling upon object representations is crucial.
We note that we compare the KL-Divergence from unigram distribution only to understand the models' behaviors, and the metric itself can not serve as an evaluation of a system's performance in grounded open vocabulary acquisition.

\vspace{-10pt}

\subsection{Addendum to Results in Multi-Class Incremental Learning}

We present additional results in Table~\ref{tab::multiclass}.

\begin{table}[!htp]
\scalebox{0.77}{
    \begin{tabular}{ccccc}
    \toprule
    \multirow{2}{*}{\# Samp.} & \multicolumn{2}{c}{Seen log G-PPL$_{\text{all}}$ ($\downarrow$)} & \multicolumn{2}{c}{Unseen log G-PPL$_{\text{all}}$ ($\downarrow$)} \\
    \cmidrule(lr){2-3} \cmidrule(lr){4-5}
                                & {\small\modelName}         & {\small\modelName}$_\textrm{w/o G}$        & {\small\modelName}         & {\small\modelName}$_\textrm{w/o G}$         \\
    \cmidrule(lr){1-1} \cmidrule(lr){2-2} \cmidrule(lr){3-3} \cmidrule(lr){4-4}  \cmidrule(lr){5-5}
    0                           & \textbf{1.79}             & 2.33              & \textbf{11.58}            & 11.89              \\
    8          & \textbf{3.15}             & 3.63              & \textbf{3.09}             & 3.32               \\
    16                          & \textbf{3.36}             & 3.76              & \textbf{2.64}             & 2.85               \\
    24                          & \textbf{3.05}             & 3.46              & \textbf{2.07}             & 2.67               \\
    32                          & \textbf{3.07}             & 3.62              & \textbf{2.01}             & 2.54              \\
    \bottomrule
    \end{tabular}
}
    \caption{The log G-PPL (All-Protocol) of seen and unseen words in multi-class incremental learning, each unseen word with a sample size ranging from 8 to 32. \vspace{-10pt}}
    \label{tab::multiclass}
\end{table}

\subsection{Learning New Words through One-Class Incremental Learning.}

We further perform a more controlled study with a word-specific one-class incremental learning setting.
The pre-trained model is tasked to acquire one single unseen word from a few-shot learning session with $|\mathcal{V}_{\text{unseen}}|=1$.
The results of this section are obtained from the test immediately following the new session.
We present the test result in Table~\ref{tab::oneclass-full}.
Again, we observe that with as few as 8 samples, {\modelName} can achieve a satisfyingly low grounded perplexity.
In the majority of the cases, {\modelName} demonstrates the better ability to acquire unseen words over the groundless baseline.

\begin{table*}[!htp]
\centering
\scalebox{0.785}{
\begin{tabular}{cccccccccccccc}
\toprule
\multicolumn{2}{c}{\# Samples}          & 0     & 8     & 16   & 24   & 32   & \multicolumn{2}{c}{\# Samples}          & 0     & 8    & 16   & 24   & 32   \\
\cmidrule(lr){1-2} \cmidrule(lr){3-7} \cmidrule(lr){8-9} \cmidrule(lr){10-14}
\multirow{2}{*}{crosswalk} & {\small\modelName}   & 10.82 & 8.48  & 7.43 & 7.70 & 5.95 & \multirow{2}{*}{donkey}    & {\small\modelName}   & 8.70  & 0.84 & 0.81 & 0.67 & 0.79 \\
                           & {\small\modelName}$_\textrm{w/o G}$ & 10.91 & 10.88 & 7.53 & 7.15 & 7.5  &                            & {\small\modelName}$_\textrm{w/o G}$ & 9.69  & 1.97 & 1.99 & 2.35 & 2.01 \\
\cmidrule(lr){1-2} \cmidrule(lr){3-7} \cmidrule(lr){8-9} \cmidrule(lr){10-14}
\multirow{2}{*}{cheese}    & {\small\modelName}   & 12.16 & 2.62  & 3.00    & 1.27 & 1.04 & \multirow{2}{*}{barefoot}  & {\small\modelName}   & 9.71  & 6.93 & 4.58 & 5.55 & 6.27 \\
                           & {\small\modelName}$_\textrm{w/o G}$ & 13.07 & 2.81  & 3.13 & 2.56 & 1.49 &                            & {\small\modelName}$_\textrm{w/o G}$ & 9.95  & 6.52 & 4.67 & 5.74 & 5.88 \\
\cmidrule(lr){1-2} \cmidrule(lr){3-7} \cmidrule(lr){8-9} \cmidrule(lr){10-14}
\multirow{2}{*}{star}      & {\small\modelName}   & 8.70  & 1.49  & 1.47 & 1.09 & 1.18 & \multirow{2}{*}{elephant}  & {\small\modelName}   & 15.24 & 1.44 & 1.65 & 1.81 & 1.44 \\
                           & {\small\modelName}$_\textrm{w/o G}$ & 10.59 & 2.93  & 2.10 & 1.99 & 1.39 &                            & {\small\modelName}$_\textrm{w/o G}$ & 14.75 & 2.17 & 1.98 & 1.73 & 1.61 \\
\cmidrule(lr){1-2} \cmidrule(lr){3-7} \cmidrule(lr){8-9} \cmidrule(lr){10-14}
\multirow{2}{*}{classroom} & {\small\modelName}   & 3.96  & 0.47  & 0.36 & 0.43 & 0.32 & \multirow{2}{*}{heart}     & {\small\modelName}   & 9.34  & 2.97 & 1.90 & 1.76 & 1.76 \\
                           & {\small\modelName}$_\textrm{w/o G}$ & 5.10  & 0.95  & 0.88 & 1.05 & 0.95 &                            & {\small\modelName}$_\textrm{w/o G}$ & 9.31  & 2.99 & 2.50 & 2.65 & 2.96 \\
\cmidrule(lr){1-2} \cmidrule(lr){3-7} \cmidrule(lr){8-9} \cmidrule(lr){10-14}
\multirow{2}{*}{fluffy}    & {\small\modelName}   & 16.44 & 1.88  & 1.78 & 0.82 & 1.36 & \multirow{2}{*}{gym}       & {\small\modelName}   & 5.13  & 2.14 & 0.44 & 0.74 & 0.69 \\
                           & {\small\modelName}$_\textrm{w/o G}$ & 15.61 & 1.83  & 1.71 & 1.37 & 1.47 &                            & {\small\modelName}$_\textrm{w/o G}$ & 4.88  & 3.73 & 1.30 & 1.08 & 1.45 \\
\cmidrule(lr){1-2} \cmidrule(lr){3-7} \cmidrule(lr){8-9} \cmidrule(lr){10-14}
\multirow{2}{*}{circular}  & {\small\modelName}   & 15.21 & 1.59  & 1.07 & 1.55 & 1.23 & \multirow{2}{*}{security}  & {\small\modelName}   & 15.08 & 1.07 & 0.81 & 1.28 & 0.71 \\
                           & {\small\modelName}$_\textrm{w/o G}$ & 15.12 & 2.25  & 2.25 & 1.81 & 1.61 &                            & {\small\modelName}$_\textrm{w/o G}$ & 14.75 & 1.50 & 1.22 & 1.53 & 1.17 \\
\cmidrule(lr){1-2} \cmidrule(lr){3-7} \cmidrule(lr){8-9} \cmidrule(lr){10-14}
\multirow{2}{*}{sink}      & {\small\modelName}   & 14.23 & 1.17  & 0.92 & 1.11 & 1.38 & \multirow{2}{*}{cafe}      & {\small\modelName}   & 6.28  & 1.90 & 1.38 & 1.98 & 1.39 \\
                           & {\small\modelName}$_\textrm{w/o G}$ & 15.49 & 1.84  & 1.65 & 1.60 & 1.84 &                            & {\small\modelName}$_\textrm{w/o G}$ & 7.03  & 2.17 & 1.92 & 2.08 & 1.72 \\
\cmidrule(lr){1-2} \cmidrule(lr){3-7} \cmidrule(lr){8-9} \cmidrule(lr){10-14}
\multirow{2}{*}{doctor}    & {\small\modelName}   & 13.03 & 1.17  & 1.05 & 1.38 & 1.18 & \multirow{2}{*}{teacher}   & {\small\modelName}   & 16.68 & 1.95 & 2.15 & 1.52 & 1.48 \\
                           & {\small\modelName}$_\textrm{w/o G}$ & 12.44 & 1.17  & 1.23 & 1.39 & 1.58 &                            & {\small\modelName}$_\textrm{w/o G}$ & 16.08 & 2.68 & 2.37 & 1.85 & 1.83 \\
\cmidrule(lr){1-2} \cmidrule(lr){3-7} \cmidrule(lr){8-9} \cmidrule(lr){10-14}
\multirow{2}{*}{foreign}   & {\small\modelName}   & 9.48  & 0.62  & 0.95 & 0.85 & 0.47 & \multirow{2}{*}{student}   & {\small\modelName}   & 16.28 & 1.38 & 1.07 & 1.20 & 1.03 \\
                           & {\small\modelName}$_\textrm{w/o G}$ & 10.01 & 1.03  & 0.88 & 1.18 & 0.95 &                            & {\small\modelName}$_\textrm{w/o G}$ & 16.52 & 2.21 & 1.29 & 1.40 & 1.61 \\
\cmidrule(lr){1-2} \cmidrule(lr){3-7} \cmidrule(lr){8-9} \cmidrule(lr){10-14}
\multirow{2}{*}{diverse}   & {\small\modelName}   & 16.44 & 0.60  & 0.22 & 0.52 & 0.24 & \multirow{2}{*}{newborn}   & {\small\modelName}   & 16.43 & 1.71 & 0.88 & 0.91 & 1.11 \\
                           & {\small\modelName}$_\textrm{w/o G}$ & 16.05 & 0.81  & 0.65 & 0.97 & 0.65 &                            & {\small\modelName}$_\textrm{w/o G}$ & 16.30 & 2.02 & 1.32 & 1.61 & 1.76 \\
\cmidrule(lr){1-2} \cmidrule(lr){3-7} \cmidrule(lr){8-9} \cmidrule(lr){10-14}
\multirow{2}{*}{product}   & {\small\modelName}   & 10.25 & 0.84  & 0.75 & 1.39 & 1.15 & \multirow{2}{*}{pan}       & {\small\modelName}   & 12.04 & 1.70 & 2.12 & 1.87 & 2.02 \\
                           & {\small\modelName}$_\textrm{w/o G}$ & 12.28 & 1.15  & 0.81 & 0.99 & 0.76 &                            & {\small\modelName}$_\textrm{w/o G}$ & 11.88 & 2.84 & 3.62 & 2.68 & 2.50 \\
\cmidrule(lr){1-2} \cmidrule(lr){3-7} \cmidrule(lr){8-9} \cmidrule(lr){10-14}
\multirow{2}{*}{stove}     & {\small\modelName}   & 16.15 & 2.63  & 2.64 & 1.94 & 2.72 & \multirow{2}{*}{telephone} & {\small\modelName}   & 14.09 & 1.18 & 0.96 & 1.05 & 0.96 \\
                           & {\small\modelName}$_\textrm{w/o G}$ & 16.13 & 3.06  & 4.30 & 3.08 & 2.98 &                            & {\small\modelName}$_\textrm{w/o G}$ & 13.42 & 1.17 & 1.50 & 1.46 & 1.38 \\
\cmidrule(lr){1-2} \cmidrule(lr){3-7} \cmidrule(lr){8-9} \cmidrule(lr){10-14}
\multirow{2}{*}{steep}     & {\small\modelName}   & 5.89  & 0.63  & 0.39 & 0.53 & 0.42 & \multirow{2}{*}{bamboo}    & {\small\modelName}   & 14.54 & 2.02 & 1.20 & 0.76 & 1.02 \\
                           & {\small\modelName}$_\textrm{w/o G}$ & 7.30  & 1.46  & 2.42 & 0.87 & 1.93 &                            & {\small\modelName}$_\textrm{w/o G}$ & 15.40  & 3.01 & 1.38 & 1.09 & 1.42 \\
\cmidrule(lr){1-2} \cmidrule(lr){3-7} \cmidrule(lr){8-9} \cmidrule(lr){10-14}
\multirow{2}{*}{warm}      & {\small\modelName}   & 7.79  & 0.68  & 0.69 & 0.68 & 0.69 & \multirow{2}{*}{brush}     & {\small\modelName}   & 11.17 & 1.88 & 2.13 & 1.81 & 2.45 \\
                           & {\small\modelName}$_\textrm{w/o G}$ & 8.67  & 1.05  & 1.01 & 0.79 & 0.85 &                            & {\small\modelName}$_\textrm{w/o G}$ & 13.69 & 2.51 & 2.89 & 2.39 & 2.83 \\
\cmidrule(lr){1-2} \cmidrule(lr){3-7} \cmidrule(lr){8-9} \cmidrule(lr){10-14}
\multirow{2}{*}{aged}      & {\small\modelName}   & 13.72  & 0.50 & 0.53 & 0.39 & 0.66 & \multirow{2}{*}{button}     & {\small\modelName}   & 4.73 & 2.37 & 2.08 & 1.82 & 2.01 \\
                           & {\small\modelName}$_\textrm{w/o G}$ & 13.50 & 0.77 & 0.94 & 0.78 & 0.93 &                            & {\small\modelName}$_\textrm{w/o G}$ & 5.94 & 3.25 & 3.19 & 2.54 & 2.74  \\
\cmidrule(lr){1-2} \cmidrule(lr){3-7} \cmidrule(lr){8-9} \cmidrule(lr){10-14}
\multirow{2}{*}{pizza}      & {\small\modelName}   & 10.70  & 1.47 & 1.07 & 1.19 & 0.90 & \multirow{2}{*}{}     & & & & & & \\
                           & {\small\modelName}$_\textrm{w/o G}$ & 9.59 & 2.21 & 2.54 & 1.25 & 1.18 &                            & & & & & &  \\
\bottomrule
\end{tabular}
}
\caption{The log G-PPL (All-Protocol) of unseen words in one-class incremental learning, each unseen word with a sample size ranging from 8 to 32. \vspace{-15pt}}
\label{tab::oneclass-full}
\end{table*}

\end{document}